\documentclass{article}

\usepackage{microtype}
\usepackage{graphicx}
\usepackage{subcaption}
\usepackage{booktabs} 

\usepackage{hyperref}

\usepackage[accepted]{icml2024}

\usepackage{amsmath}
\usepackage{amssymb}
\usepackage{mathtools}
\usepackage{amsthm}

\usepackage{ulem}

\usepackage[capitalize,noabbrev]{cleveref}

\theoremstyle{plain}

\theoremstyle{definition}

\theoremstyle{remark}

\usepackage[textsize=tiny]{todonotes}

\icmltitlerunning{WISER: Weak supervISion and supErvised Representation learning to improve drug response prediction in cancer}

\usepackage{amsmath,amsfonts,bm}

\def\eqref#1{equation~\ref{#1}}

\def\1{\bm{1}}

\def\vone{{\bm{1}}}

\def\vl{{\bm{l}}}

\def\vs{{\bm{s}}}
\def\vt{{\bm{t}}}

\def\vw{{\bm{w}}}

\def\vy{{\bm{y}}}
\def\vz{{\bm{z}}}

\DeclareMathAlphabet{\mathsfit}{\encodingdefault}{\sfdefault}{m}{sl}
\SetMathAlphabet{\mathsfit}{bold}{\encodingdefault}{\sfdefault}{bx}{n}

\def\gB{{\mathcal{B}}}
\def\gC{{\mathcal{C}}}
\def\gD{{\mathcal{D}}}
\def\gE{{\mathcal{E}}}
\def\gF{{\mathcal{F}}}
\def\gG{{\mathcal{G}}}

\def\gJ{{\mathcal{J}}}

\def\gL{{\mathcal{L}}}
\def\gM{{\mathcal{M}}}
\def\gN{{\mathcal{N}}}
\def\gO{{\mathcal{O}}}
\def\gP{{\mathcal{P}}}

\def\gR{{\mathcal{R}}}
\def\gS{{\mathcal{S}}}

\def\gV{{\mathcal{V}}}
\def\gW{{\mathcal{W}}}
\def\gX{{\mathcal{X}}}
\def\gY{{\mathcal{Y}}}
\def\gZ{{\mathcal{Z}}}

\begin{document}

\twocolumn[
\icmltitle{WISER: Weak supervISion and supErvised Representation learning to improve drug response prediction in cancer}



\icmlsetsymbol{equal}{*}

\begin{icmlauthorlist}
\icmlauthor{Kumar Shubham}{1}
\icmlauthor{Aishwarya Jayagopal}{3}
\icmlauthor{ Syed Mohammed Danish}{2}
\icmlauthor{Prathosh AP}{1}
\icmlauthor{Vaibhav Rajan}{3}
\end{icmlauthorlist}

\icmlaffiliation{1}{Department of Electrical Communication Engineering, Indian Institute of Science, Bengaluru, Karnataka, India}
\icmlaffiliation{2}{Indian Institute of Technology, Patna, Bihar, India}
\icmlaffiliation{3}{Department of Information Systems and Analytics, National University of Singapore, Kent Ridge, Singapore}
\icmlcorrespondingauthor{Kumar Shubham}{shubhamkuma3@iisc.ac.in}
\icmlcorrespondingauthor{Vaibhav Rajan}{vaibhav.rajan@nus.edu.sg}


\vskip 0.3in
]



\printAffiliationsAndNotice{}  

\begin{abstract}
Cancer, a leading cause of death globally, occurs due to genomic changes and manifests heterogeneously across patients.
To advance research on personalized treatment strategies, the effectiveness of various drugs on cells derived from cancers (`cell lines') is experimentally determined in laboratory settings. Nevertheless, variations in the distribution of genomic data and drug responses between cell lines and humans arise due to biological and environmental differences. Moreover, while genomic profiles of many cancer patients are readily available, the scarcity of corresponding drug response data limits the ability to train machine learning models that can predict drug response in patients effectively.
Recent cancer drug response prediction methods have largely followed the paradigm of unsupervised domain-invariant representation learning followed by a downstream drug response classification step.
Introducing supervision in both stages is challenging due to heterogeneous patient response to drugs and limited drug response data. This paper addresses these challenges through a novel representation learning method in the first phase and weak supervision in the second. Experimental results on real patient data demonstrate the efficacy of our method (WISER) over state-of-the-art alternatives on predicting personalized drug response. Our implementation is
available at \href{https://github.com/kyrs/WISER}{https://github.com/kyrs/WISER}
\end{abstract}

\section{Introduction}
\label{sec:intro}
Cancer is a major cause of global morbidity and mortality \cite{who}.
Cancer develops due to changes in our genome, which enable cancer cells to gain a selective advantage over healthy cells, resulting in uncontrolled proliferation as a cancerous tumour. 
Significant variability in treatment sensitivity and outcomes among patients makes cancer treatment difficult \cite{bedard2013tumour}. 
Hence, cancer care is transitioning from a `one-size-fits-all' approach to a more personalized strategy guided by patient-specific genomic characteristics \cite{wahida2023coming}.

To aid therapeutic development, there have been large-scale global efforts, e.g., through The Cancer Genome Atlas (TCGA) database \cite{tcga},  
to catalog high-dimensional genomic information ($\gX$) of cancer patients. 
However, patient drug response data 
\big[$\gY^{d_i}_t (\gX)$ for drug $d_i$\big]
is scarce due to limited number of patients, with only a few drugs administered on each patient~\cite{sharifi2021out}. This has motivated researchers to explore   
preclinical datasets -- e.g., \textit{cell lines}, comprising cells extracted from patient cancers, which can be cloned in a way that the same genomic information is replicated across them.
Such clones can be exposed to different drugs to obtain drug response information $\gY^{d_i}_c (\gX)$ for multiple $d_i$ on the same $\gX$.
This data is immensely useful and cannot be directly obtained from patients, who cannot be subjected to multiple drug regimens simultaneously. 
While such fine-grained drug response data is only available for a limited number of cell lines ($\sim$ 1000) and drugs, it provides a valuable starting point to build personalized drug response models based on genomic information.

However, previous studies have shown that such cell line-based response models do not accurately predict drug efficacy in patients due to several reasons \cite{seyhan2019lost}.
Cell line data ($\gX_c$) is more homogeneous than patient cancer cells ($\gX_t$) and the environments in which they reside are different. 
This results in differences in the distributions ($P$) of genomic information across cell lines and patients $\big(P(\gX_c) \neq P(\gX_t)\big)$, and they can be considered as different domains (See Appendix~\ref{appx: domain_details}).
Further, within the human body, in addition to the genomic structure, several other factors (e.g., the immune system) play a role in drug response.
Thus, the drug response functions are different across cell lines and patients 
\big(i.e., $\gY^{d_i}_t (.) \neq \gY^{d_i}_c (.)$\big).

To address these challenges, several domain adaptation and transfer learning-based drug response models, that use a combination of cell line and patient data, have been developed. These methods generally consist of two stages: (1) an unsupervised representation learning phase where domain-invariant representations of genomic data are learned and, (2) a classification phase where these representations are used to train a drug response prediction model by categorizing responses as positive or negative based on the drug's impact on inhibiting cancer growth. 
The classifier is trained using labeled data and used to predict drug response in patients.


Unsupervised representation learning approaches used by extant methods do not consider the drug response information \big($\gY^{d_i}_c (\gX_c)$\big) associated with genomic profiles in cell lines, and hence do not  distinguish between responders and non-responders to drugs. Supervised contrastive learning approaches~\cite{unbiased_sup_con,sup_con,graf2021dissecting,triplet_loss_defence,triplet_loss, lee2021attention} can address this by bringing the representations ($\gZ$) of data points with similar class labels closer together i.e., $\gZ(\gX^m_c) \sim  \gZ(\gX^n_c)$ if $\gY^{d_i}_c(\gX^m_c)= \gY^{d_i}_c(\gX^n_c)$, emphasizing their shared characteristics over dissimilar classes. However, genomic profiles that respond to one drug may behave differently for another i.e, for drug $d_i$, $\gY^{d_i}_c(\gX^m_c) = \gY^{d_i}_c(\gX^n_c)$ but for drug $d_k$, $\gY^{d_k}_c(\gX^m_c) \neq \gY^{d_k}_c(\gX^n_c)$. Hence, for drug $d_i$, $\gZ(\gX^m_c) \sim  \gZ(\gX^n_c)$  but for drug $d_k$, $\gZ(\gX^m_c) \nsim  \gZ(\gX^n_c)$. 
Further, limited patient data with documented drug response makes it difficult to find genomic profiles with similar efficacy across multiple drugs. These difficulties, in turn, limit the ability to use standard supervised contrastive learning methods to bring the representations of genomic profiles closer together. Our study addresses this challenge by learning a discrete representation per drug ($\gR$)~\cite{VQVAE} and representing each genomic profile as a weighted combination ($\gZ = \sum\gW\gR$). 
To ensure that $\gZ$ is simultaneously reflective of the responses from multiple drugs, we increase the weights of drugs with positive response compared to those with negative response, through a supervised triplet loss~\cite{triplet_loss_defence,triplet_loss,unbiased_sup_con}.

It is worth noting that while there is scarcity of labeled data in both domains, relatively abundant unlabeled patient data is available (See Appendix~\ref{appx: domain_details}). While prior studies have leveraged unlabeled patient data for learning domain-invariant representations, the training of drug response prediction classifiers has predominantly relied on the cell line dataset due to insufficient labeled response data for patients. Techniques like weak supervision can be employed to generate pseudo-labels for the abundant 
unlabeled data. However, na\"ively using all pseudo-labeled samples does not improve performance~\cite{subset_ws,shubham2023fusing} (also seen in our results). In fact, there exists a trade-off between the noise introduced in the downstream classifier due to pseudo labels and the generalization it achieves when trained in a weak supervision setting~\cite{subset_ws}. To address this, we introduce a subset selection step~\cite{subset_ws, shubham2023fusing}, which to our knowledge is novel in this context and helps boost performance.  
We employ majority-vote-based weak supervision techniques~\cite{ratner2017snorkel,PWS_survey} to create pseudo labels for patient genomic profiles without documented drug response, followed by a subset selection strategy~\cite{cut_stats}. This subset is combined with labeled cell line data to train the drug response prediction classifier.
Our contributions can be summarized as follows:
\begin{itemize}
    \item We design a new supervised domain-invariant representation learning approach which offers better distinction between drug responders and non-responders by addressing the challenges of limited sample size and heterogeneous drug response of genomic profiles.
    \item We propose a novel strategy that carefully selects a subset of least noisy pseudo-labeled patient data for classifier training on the domain-invariant representations. 
    \item Using these techniques we propose a new method, called WISER, to estimate drug response for patients using unlabeled patient data and a small set of labeled cell line data.
    \item Our experiments on benchmark datasets demonstrate the superiority of WISER over state-of-the-art methods for drug response prediction, with improvements of up to 15.7\% in AUROC.
    \item The most important features (genes) responsible for pseudo-labeling patient samples in the selected subsets correlates well with independent clinical evidence based on gene-drug interactions that impact patient survival, which further validates our pseudo-labeling approach.
\end{itemize}



\section{Related Work}

\subsection{Drug Response Prediction}
Prior literature on drug response prediction in patients has primarily focused on transfer learning~\cite{pan2009survey}.
These approaches are useful when the target domain (patients) has limited samples, and a related source domain (cell lines) has more labeled samples.
Transductive transfer learning methods~\cite{DSN, sharifi2021out, CORAL} use labeled source domain samples for drug response prediction but often build one model per drug and, thus, lack correlations across drugs.  Inductive transfer learning methods~\cite{sharifi2020aitl, ma2021few} utilize few-shot and multi-task learning on available labeled patient samples but exhibit inferior performance to other approaches. Recent methods, like CODE-AE~\cite{code-AE} learn shared representations using unlabeled genomic profiles from both domains. 

Among the extant approaches, CODE-AE \cite{code-AE} has demonstrated superior predictive accuracy and robustness through extensive benchmark studies.
CODE-AE is trained in two stages: (1) unsupervised pretraining of autoencoders to learn both domain-specific private and domain-invariant shared representations and (2) downstream drug response prediction based on the learned shared representations. 
A key shortcoming of this approach is that 
the representations learnt do not factor in the downstream drug response prediction task. Further, they do not utilize a large number of unlabeled patient genomic profiles in the downstream drug response prediction. Our proposed method, WISER, can handle these shortcomings and differs from CODE-AE in two aspects - (1) we incorporate drug response information of cell lines through supervised domain-invariant representation learning, and (2) we also utilize the available unlabeled patient genomic profiles through weak supervision techniques followed by subset selection. 
%

\subsection{Weak Supervision and Subset Selection}

Weak supervision techniques~\cite{PWS_ratner2016} are designed to address the challenge of limited data size. They leverage information from various sources (\textit{label functions}), such as data from different domains~\cite{mazzetto2021_domainnet, PWS_survey}, to generate cost-effective but noisy labels for unlabeled data.
To further enhance the accuracy of the estimation process, confident predictions from different sources of pseudo labels are systematically combined, through weighing or voting schemes~\cite{PWS_ratner2016,PWS_dawid1979,PWS_fu_fastsquid}. 
For a smaller set of label functions, a majority vote-based scheme~\cite{ratner2017snorkel} outperforms weighing techniques.

In addition, recent works~\cite{subset_ws,shubham2023fusing}, in weak supervision, have shown that a subset of original data can generate optimal results compared to the use of the entire pseudo-labeled dataset. In fact, there exists a trade-off between the generalization achieved by the classifier and the noise introduced by the pseudo labels. 
Previous studies on subset selection have primarily concentrated on natural language tasks and employed pre-trained word embeddings~\cite{kenton2019bert}. However, the application of these on cancer research remains unexplored.

\section{Proposed Method}
\subsection{Problem Formulation and Solution Overview}
\textbf{\textit{Problem Definition:}} Let us assume that there are $\gN_{c}$ labeled samples of genomic profiles associated with cell lines \big($\gG_{cell}$\big) and $\gN_{t}$ unlabeled samples of genomic profiles from patients \big($\gG_{patient}$\big).
In this work, although we focus on gene expression profiles, our method can also be applied to other omics data types, such as mutations.
In general, $\gN_{c} <<\gN_{t}$.
Let $\{d_1,d_2 \hdots d_n\}$ be the set of $n$ drugs with documented drug response for $\gG_{cell}$ and $\gY^{d_i}_c \big(\gX^j_c \big)$ be the  corresponding response of a drug \big($d_i$\big) to a genomic profile $\gX^j_c \in \gG_{cell}$ and $\gY^{d_i}_t \big(\gX^m_t \big)$ be the drug response for patients $\gX^m_t \in \gG_{patient}$. Note that $\gY^{d_i}_c \big(\gX^j_c\big) \in \{ -1, 0,  1\}$  where 1 indicates a positive response of a genomic profile $\gX^j_c$ to drug $d_{i}$, 0 indicates a negative response to the drug $d_{i}$ and -1 represents that the response data is not available. The main objective of our work is to use the labeled cell line data \big($\gG_{cell}, \gY^{d_i}_c$\big) for n drugs $\{d_1,d_2 \hdots d_n\}$ and the unlabeled patient genomic profile \big($\gG_{patient}$\big) to estimate drug response for patients \big($\gY^{d_i}_t$\big). Further details about both domains are provided in Appendix~\ref{appx: domain_details}.

\begin{figure*}[!t]
\centering
\begin{tabular}{@{}c|c@{}}
\begin{subfigure}[t]{0.38\textwidth}
    \includegraphics[width=\textwidth]{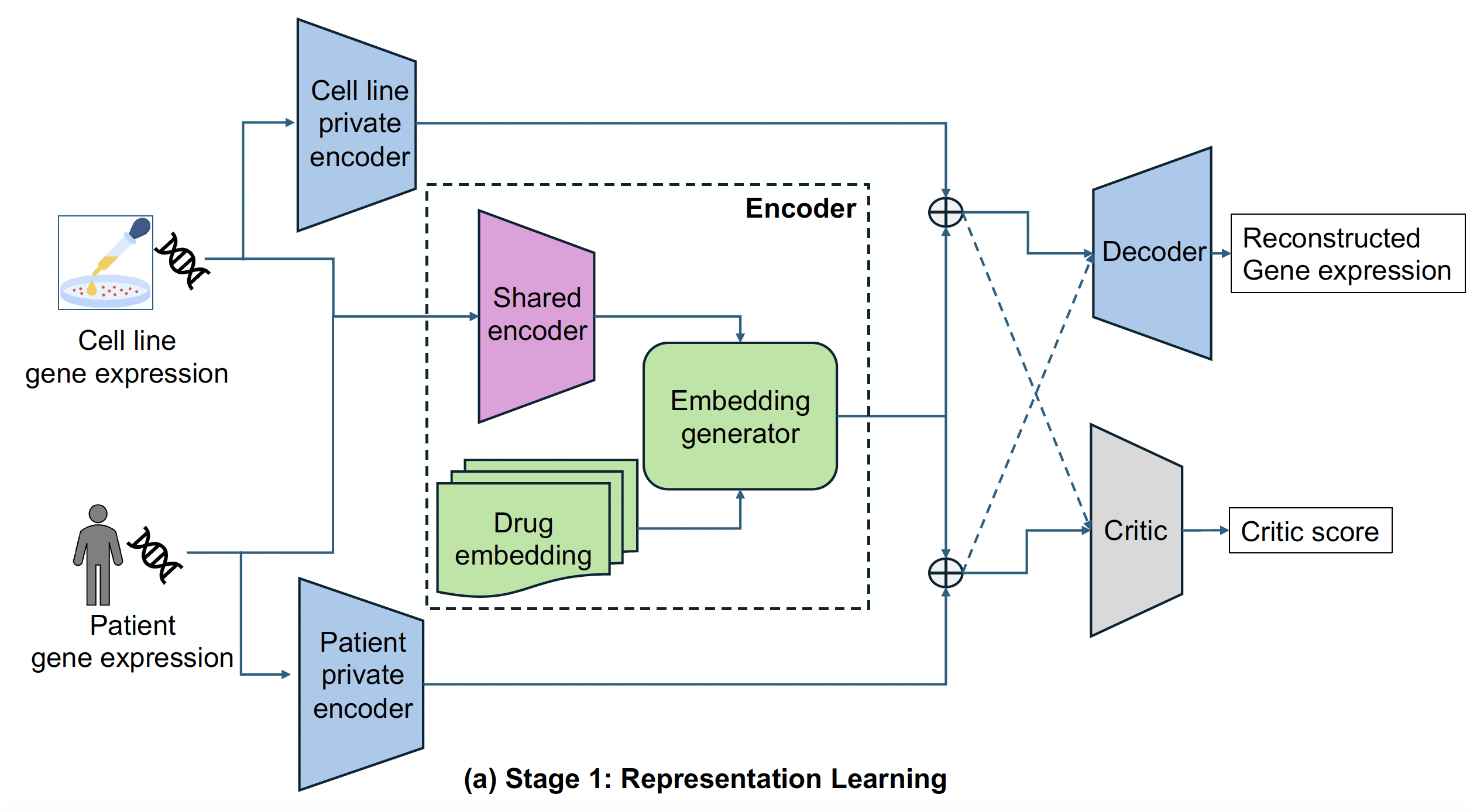}
\end{subfigure}
&
\begin{subfigure}[t]{0.38\textwidth}
    \includegraphics[width=\textwidth]{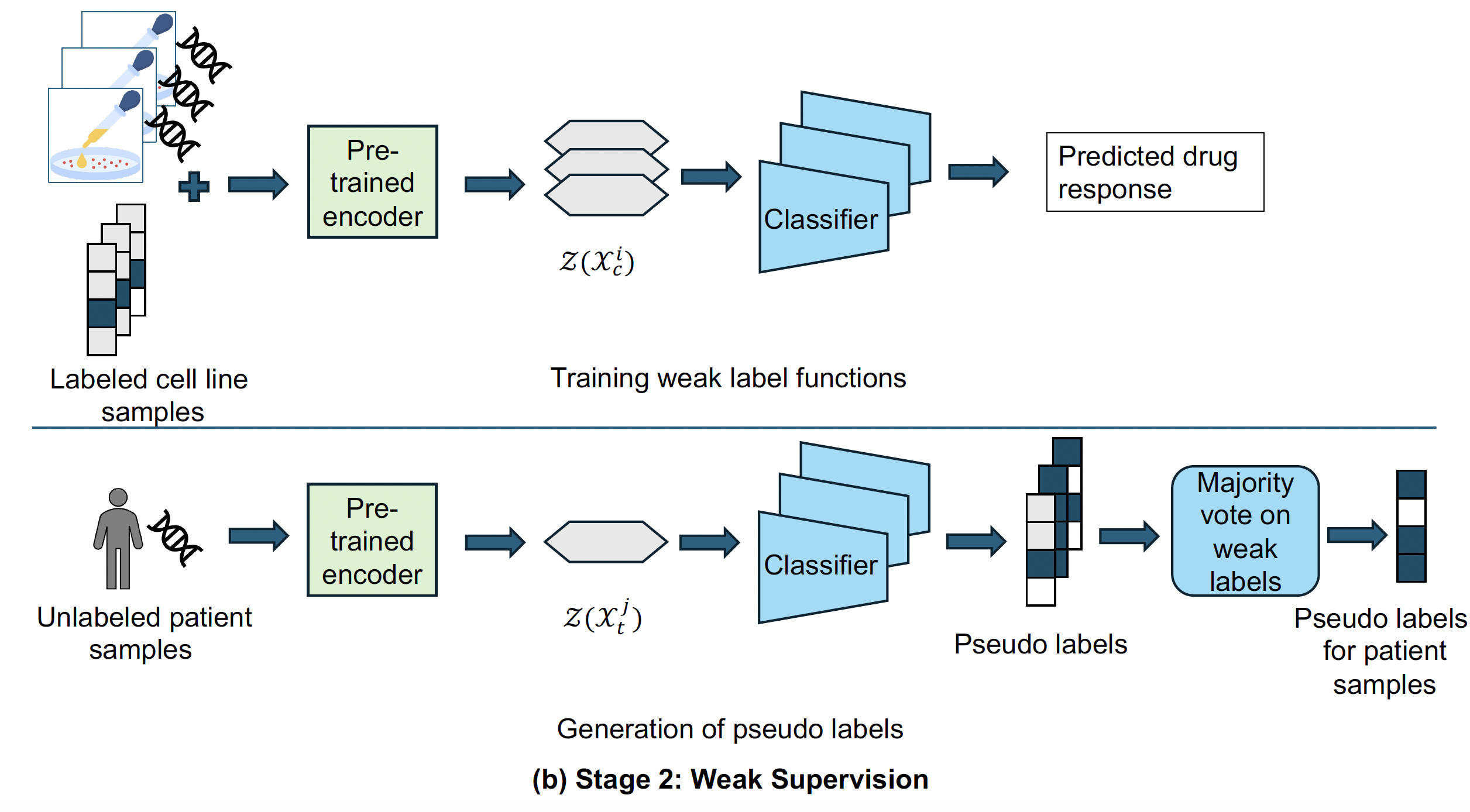}
\end{subfigure}
\\
& \\
\hline
& \\
\begin{subfigure}{0.38\textwidth}
    \includegraphics[width=\textwidth]{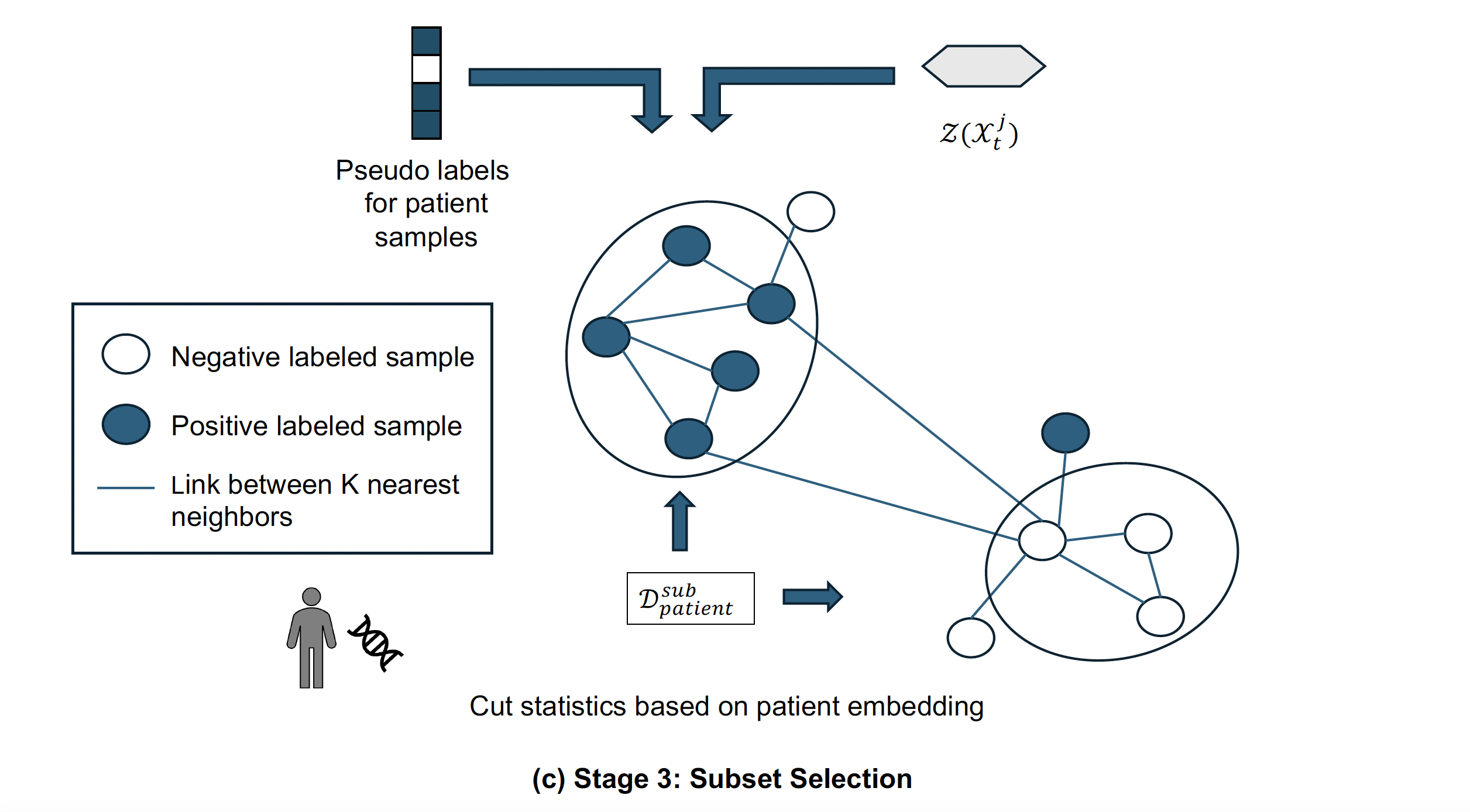}
\end{subfigure}
&
\begin{subfigure}{0.38\textwidth}
    \includegraphics[width=\textwidth]{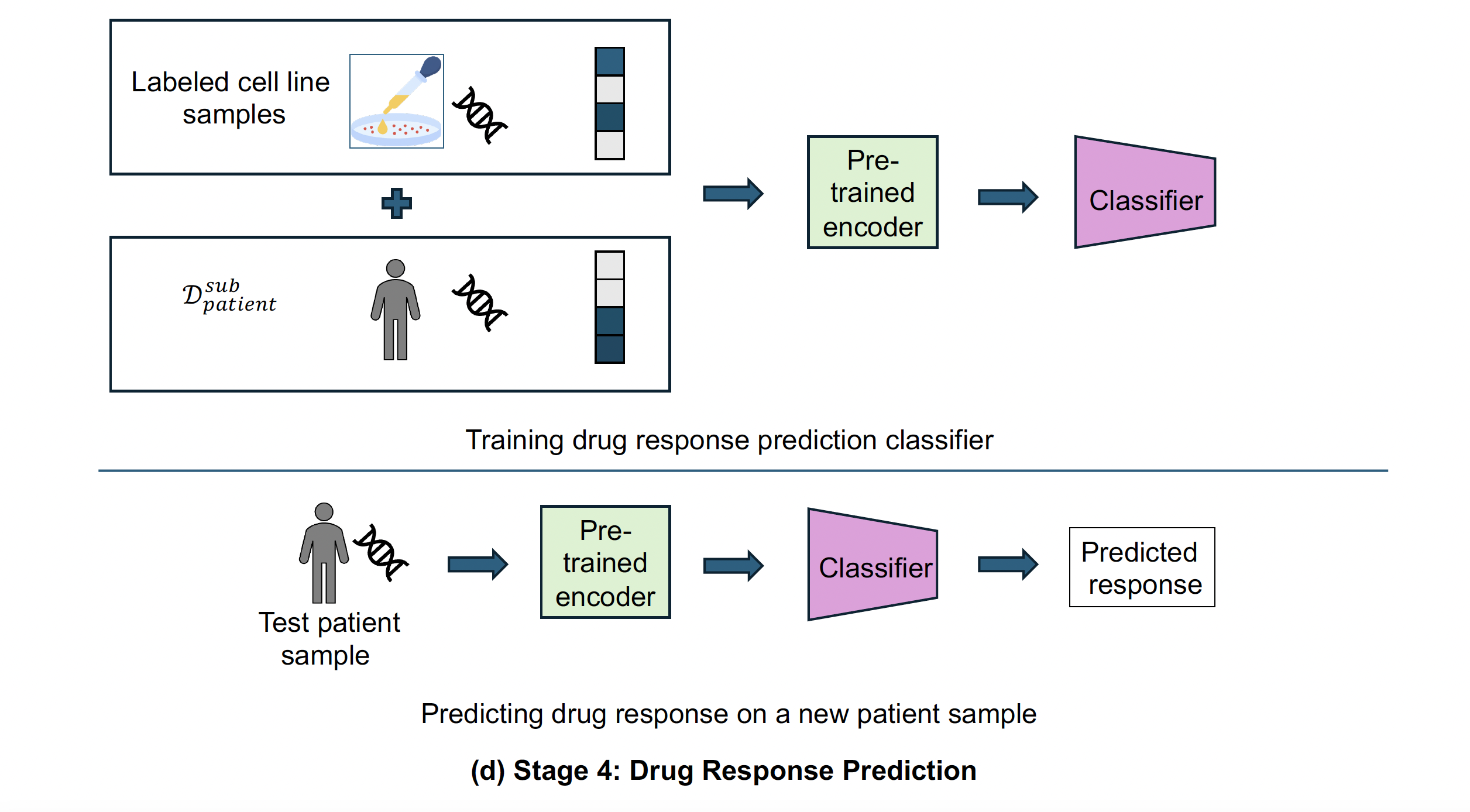}
\end{subfigure}
\end{tabular}
\caption{This diagram outlines WISER's comprehensive training process, divided into four key phases. First, in the Representation Learning phase, a domain-invariant representation ($\gZ$) is learned between cell line and patient genomic profiles using a shared encoder and private encoding scheme. Next, in the Weak Supervision phase, multiple label functions are trained using labeled genomic profiles of cell lines to assign pseudo labels to unlabeled patient genomic profiles. Following that, in the Subset Selection phase, pseudo labels and the domain-invariant representation ($\gZ$) are used to select a subset of patient genomic profiles ($\gD_{patient}^{sub}$) and associated pseudo labels based on the consistency of the labels among nearest neighbors. Finally, in the Drug Response Prediction phase, the selected subset, along with labeled genomic profiles from cell lines, is utilized for downstream classifier training and predicting drug responses among patients. }
\label{fig:architecture}
\vspace{-0.5cm}
\end{figure*}

\textbf{\textit{Solution Overview:}} 
Here, we describe the overview of our method comprising four major stages as depicted in Fig. \ref{fig:architecture}.  
 
\textbf{Stage 1: Representation Learning} In the first stage, we learn representations that are invariant between patient and cell line domains. 
Specifically, we learn discrete latent representations for individual drugs.  The desired domain-invariant representation $\gZ$ is generated through a weighted combination of these drug representations. 


\textbf{Stage 2: Weak Supervision} To incorporate the unlabeled patient genomic profiles in the training of the downstream drug response prediction model, we train multiple classifiers (\textit{label functions}) using labeled cell line data and the domain invariant representation \big($\gZ$\big). These label functions are then used to predict labels for the unlabeled patient dataset. The confident predictions from all label functions are combined based on majority-vote to assign the pseudo labels. 

\textbf{Stage 3: Subset Selection} In this stage, we propose to utilize a subset of genomic profiles with confident predictions as indicated by the label functions. We employ cut statistics~\cite{cut_stats} in conjunction with the domain-invariant representation \big($\gZ$\big) to select a subset of least noisy samples.


\textbf{Stage 4: Drug Response Prediction} We combine the subset of patient genomic profiles and associated pseudo labels, chosen after subset selection in Stage 3, with the labeled cell line genomic profiles to train a downstream drug response prediction classifier. This classifier can be used to infer drug responses in new patients.




\subsection{Representation Learning}
Genomic profile data collected from cell lines and patients exhibit distributional shifts owing to multiple confounding factors~\cite{code-AE}. This can cause a model trained using cell line data to not generalize to patients. In line with previous work, we address this using a private and shared encoder scheme, where a shared encoder \big($\gC_{\gS}$\big) captures a domain invariant representation between the two domains while a private encoder \big($\gC_{\gP}$\big) captures domain specific information. 
However, \citet{code-AE} do not consider the drug response information \big($ \gY^{d_i}_c(\gX_c) $\big) during representation learning. 
We address this by representing the genomic profile ($\gZ$) as a weighted combination of drug embedding ($\gR$) (Eq.~\ref{eq:z_repr}, Eq.~\ref{eq:vqvae}) and used a triplet loss to learn these weights based on the drug efficacy results (Eq.~\ref{eqn:cns_loss}). 

In line with discrete representation learning methods~\cite{lee2021attention}, we leverage information on how a specific drug responds to a genomic profile, to generate a drug-specific discrete latent representation \big($\gR = \{\gR_{d_1}, \gR_{d_2}  \hdots \gR_{d_n} \}$\big). Similarly, inspired by contextual attention maps~\cite{NTM,bahdanau2014neural}, we combine the discrete representations of drugs \big($\gR$\big) and the shared representation of genomic profiles \big($\gC_{\gS}$\big) to form a new representation of the given genomic profile \big($\gZ$\big). This new representation is a weighted sum of drug embeddings \big($\gR$\big), with weights \big($\gW$\big) indicating the efficacy of the different drugs on a given genomic profile. To obtain $\gW$, we calculate the cosine similarity \big($sim(.)$\big) between $\gR$ and $\gC_{\gS}$. 

The scores over different drugs are further normalized using a softmax function with an inverse temperature \big($\Delta$\big) to generate the weight $\gW$. A weighted combination of $\gR$ using $\gW$ is used to generate $\gZ$, as given in Eq.~\ref{eq:z_repr}.
\begin{align}
\gZ(\gX) &= \sum\limits_{i=1}^{n}\gW_i\big(\gX\big)\gR_{d_i} \nonumber \\
\gW_i\big(\gX\big) &= \frac{exp\big(\Delta * sim\big(\gC_{\gS}\big(\gX\big), \gR_{d_i}\big)\big)}{\sum\limits_{j=1}^n exp\big(\Delta* sim\big(\gC_{\gS}\big(\gX\big), \gR_{d_j}\big)\big)} \nonumber \\
sim\big(\gC_{\gS}\big(\gX\big), \gR_{d_i}\big) &=  \frac{\gC_{\gS}\big(\gX\big) ^T \gR_{d_i}} { ||\gC_{\gS}\big(\gX\big)|| \text{  } ||\gR_{d_i}||}
\label{eq:z_repr}
\end{align}

For the training of our encoder models \big($\gC_s$, $\gC_{p}$\big) we concatenate ($\oplus$) the weighted representation \big($\gZ$ from Eq.\ref{eq:z_repr}\big) and the private representation \big($\gC_{\gP}  = \big \{\gC_{\gP}^t, \gC_{\gP}^c \big \}$\big) of a genomic profile before passing it through a shared decoder \big($D$\big) for reconstruction \big($\vl_1$\big) in both the domains \big($\gX_c^i \in \gG_{cell}, \gX_t^j \in \gG_{patient} $\big), with the following reconstruction loss:
\vspace{- 15 pt}
\begin{align}
    \vl_{recon} &=\frac{\sum\limits_{i=1}^{\gN_c} \vl_1\big(\gX_c^i, \gC_{\gP}^c\big) }{\gN_c} + \frac{\sum\limits_{j=1}^{\gN_t} \vl_1\big(\gX_t^j, \gC_{\gP}^t\big) }{\gN_t} \nonumber \\
   \text{Where, } \vl_1(\gX, \gC_{\gP}) &= ||D \big(\gZ(\gX) \oplus \gC_{\gP}(\gX)\big) - \gX||^2   
    \label{eqn:recon}
\end{align}

To ensure that generated embedding \big($\gZ$\big) and the private embedding \big($\gC_{\gP}$\big) do not capture redundant information, we introduce an orthogonal loss~\cite{code-AE} between these two embeddings as: $ \vl_{ortho} = ||\gZ \big(\gX_c \big)^T\gC_{\gP}^c \big(\gX_c \big)||^2 + ||\gZ \big(\gX_t \big)^T\gC_{\gP}^t \big(\gX_t \big)||^2$.

We further use the embedding loss $\vl_{embed}$~\cite{VQVAE} to ensure that the generated embedding \big($\gZ$\big) and the encoded genomic profiles \big($\gC_{\gS}$\big) are closer to each other for both cell lines and patients. Eq.~\ref{eq:vqvae} illustrates this where $sg(.)$ denotes the stop gradient operator.
\begin{align}
\vl_{embed} =\frac{\sum\limits_{i=1}^{\gN_c} \vl\big(\gX_c^i\big)}{\gN_c} + \frac{\sum\limits_{j=1}^{\gN_t} \vl\big(\gX_t^j\big)}{\gN_t}& \nonumber \\
\vl(\gX)  = ||\gZ\big(\gX\big)-sg(\gC_{\gS}\big(\gX\big)\big)||^2 \nonumber  \\+||sg\big(\gZ\big(\gX\big)\big)-\gC_{\gS}\big(\gX\big)||^2
\label{eq:vqvae}
\end{align}
To ensure that the learnt representations reflect the drug efficacy on labeled genomic profiles, we rely on supervised triplet loss~\cite{triplet_loss_defence,triplet_loss}, which has a direct correspondence to modern supervised contrastive loss~\cite{unbiased_sup_con}. Triplet loss minimizes the distance between an anchor and positive labeled samples while maximizing the distance from negative labeled samples. 
In our formulation we use the cosine distance \big($dis(\cdot ) = 1-sim(\cdot )$ in Eq. \ref{eq:z_repr}\big), with the drug representation \big($\gR$\big) as the anchor. The goal is to minimize ($\l_{cns}$) the average distance of this anchor from the genomic representation with positive efficacy \big($\vs^{+}$ \big) and maximize its distance from the genomic representation with negative efficacy \big($\vs^{-}$\big), (Eq. ~\ref{eqn:cns_loss}) where $\vone\big(\gY^{d_j}_c\big(\gX_j^{j}\big) = 1\big)$ is an indicator function capturing the positive efficacy of drug \big($d_j$\big) on genomic profile \big($\gX_j^{j}$\big) and $\vone\big(\gY^{d_j}_c\big(\gX_j^{j}\big) = 0\big)$ captures the negative efficacy; $\delta$ is the minimum offset between $\vs^{+}$ and $\vs^{-}$.
\begin{align}
\small
\vl_{cns} &= \max\big(\vs^{+}-\vs^{-} + \delta, 0\big) \nonumber \\
\vs^{+} &= \frac{\sum\limits_{i=1}^{{\gN}_{c}} \sum\limits_{j=1}^n \vone\big(\gY^{d_j}_c\big(\gX_c^{i}\big)=1\big) dis\big(\gC_s\big(\gX^{i}_c\big),\gR_{d_j}\big)}{ \sum\limits_{i=1}^{\gN_{c}} \sum\limits_{j=1}^{n} \vone\big(\gY^{d_j}_c\big(\gX_c^{i}\big) = 1\big)}  \nonumber\\ 
\vs^{-} &= \frac{\sum\limits_{i=1}^{\gN_{c}} \sum\limits_{j=1}^{n} \vone\big(\gY^{d_j}_c\big(\gX_c^{i}\big)=0\big) dis\big(\gC_s\big(\gX^{i}_c\big),\gR_{d_j}\big)}{ \sum\limits_{i=1}^{\gN_{c}} \sum\limits_{j=1}^{n} \vone\big(\gY^{d_j}_c\big(\gX_c^{i}\big)=0\big)}
\label{eqn:cns_loss}
\end{align}
\subsubsection{Domain Adaptation}
For efficient generalization of the downstream model, the generated representation should be invariant to the domain. Recent works~\cite{code-AE,ganin2016domain}, have tried to generate such a representation using adversarial networks. Within this framework, a separate critic network is trained to distinguish between the embeddings from the two domains, while an encoder tries to generate indistinguishable embeddings for the critic. This additional training step ensures that as the training proceeds, an equilibrium is reached where the embedding is invariant for the critic network \big($\gF$\big). In our work, we have used the Wasserstein GAN (WSGAN)~\cite{wsgan} with a gradient penalty-based adversarial loss~\cite{penalty-WSGAN} to train our critic network (Eq.~\ref{eqn:domain}). The critic network takes as input a concatenation \big($\hat{C}$\big) of generated embedding and private representation from both the domains. $\vl_{critic}$ tries to minimize the difference between the mean critic scores for patients $\gF\big(\hat{C}\big(\gX_t^j, \gC_{\gP}^t\big)\big)$ and cell lines $\gF\big(\hat{C}\big(\gX_c^i, \gC_{\gP}^c\big)\big)$. In contrast, the patient representations are learnt to obtain a higher critic score \big($\vl_{gen}$\big). A gradient penalty term is added, which encourages the gradient of the critic to have a norm close to 1 to maintain Lipschitz continuity~\cite{wsgan}. These gradients are calculated on linear interpolate of input representation from both the domains \big($\gL$\big), where $\gL = \epsilon \hat{C}\big(\gX_c, \gC_{\gP}^c\big) + \big(1-\epsilon\big)\hat{C}\big(\gX_t, \gC_{\gP}^t\big)$ and  $ \epsilon \sim U\big(0,1\big)$. Mathematically, the aforementioned loss functions are defined a
s follows:  
\begin{align}
\vl_{critic} &=  \frac{1}{\gN_t} \sum\limits_{j=1}^{\gN_t}\gF\big(\hat{C}\big(\gX_t^j, \gC_{\gP}^t\big)\big) \nonumber \\&- \frac{1}{\gN_{c}} \sum\limits_{i=1}^{\gN_c}\gF\big(\hat{C}\big(\gX_c^i, \gC_{\gP}^c\big)\big) \nonumber \\&+
\lambda\big(||\nabla_{\gL}\gF\big(\gL\big)|| -1\big)^2 \nonumber \\
\vl_{gen} &=   -\frac{1}{\gN_{t}} \sum\limits_{i=1}^{\gN_t}\gF\big(\hat{C}\big(\gX_t^i, \gC_{\gP}^t\big)\big) \nonumber\\
\hat{C}\big(\gX, \gC_{\gP}\big) &=\gZ\big(\gX\big) \oplus \gC_{\gP} \big(\gX\big)
\label{eqn:domain}
\end{align}

The complete training occurs in two stages - first where the model is trained only using the loss \big($\vl_{pl}= \vl_{recon} +  \vl_{cns} + \vl_{embed} +  \vl_{ortho}$ \big) for a few epochs and later using $\vl_{total} = \vl_{pl} + \vl_{gen} $, and $\vl_{critic}$ for the critic network.

\subsection{Weak Supervision}
Once we learn the domain invariant representations, they are subsequently employed to generate pseudo labels for the unlabeled genomic profile of patients. For this task, we partition the labeled cell line data into $\gO$ distinct subsets \big($\gD_{cell}^i \text{ }i \in{1\hdots \gO}$, where $\gD_{cell}^i \subset \gG_{cell}$\big) and train a classifier \big($\gM_i$\big) using their representations \big($\gZ$\big). Each individual classifier acts as a label function in our weak supervision framework and is utilized to infer the probability of drug response prediction for the genomic profile of patients \big($\gP_{i}\big(y|\gX_{t}^j\big)$, where $\gX_{t}^{j} \in \gG_{patient}$\big).
The model assigns a label $\hat{\vy} = 1$, when the predicted drug response probability exceeds a threshold $\vt^+$ and $\hat{\vy} = 0$, when the probability falls below a threshold $\vt^-$.
For all intermediate probabilities where the confidence in model predictions is low, it abstains from assigning any class and labels the sample as -1 \big(Eq \ref{eqn:WS}\big).
\begin{align}
    \gP_i\big(y|\gX_t^j\big) =&\gM_i\big(\gZ\big(\gX_t^j\big)\big) \text{ s.t. } i\in \{1 \hdots \gO\},  \gX_t^j \in \gG_{patient} \nonumber \\
\hat{\vy}^{j}_i= &
        \begin{cases}
            1,& \text{if } \gP_i\big(y|\gX_t^j\big) > t^+\\
            0,& \text{if } \gP_i\big(y|\gX_t^j\big) < t^-\\
            -1,& \text{otherwise}\\
        \end{cases}
    \label{eqn:WS}
\end{align}
Samples with atleast one valid prediction (not abstained) from the label functions are used subsequently. The final pseudo label \big($\vy^j_t$\big) for a given patient genomic profile \big($\gX_t^j$\big) is decided by a majority vote across all non abstained predictions \big($\hat{\vy}^{j}_i$\big). The details are in Eq.~\ref{eqn:MV}, where \big($\vone\big(\hat{\vy}^{j}_i=1\big)$\big) and \big($\vone\big(\hat{\vy}^{j}_i=0\big)$\big) are indicator functions. 
\begin{align}
    \vy^j_t=& 
    \begin{cases}
    1,&  \text{if }  
\sum\limits_{i=1}^{\gO}\vone\big(\hat{\vy}^{j}_i=1\big)  >  \sum\limits_{i=1}^{\gO} \vone\big(\hat{\vy}^{j}_i=0\big) \\
    0, & \text{if} \sum\limits_{i=1}^{\gO}\vone\big(\hat{\vy}^{j}_i=1\big)  \leq  \sum\limits_{i=1}^{\gO} \vone\big(\hat{\vy}^{j}_i=0\big)
    \end{cases}
    \label{eqn:MV}
\end{align}



\subsection{Subset selection and Drug Response Prediction}
Once the pseudo labels have been assigned to the non-abstained patient genomic profiles, they can be directly used in conjunction with the labeled cell line data for the training of the drug response prediction classifier. However, recent works~\cite{subset_ws,shubham2023fusing} have shown that in a weak supervision setting, a complete set of non-abstained samples generates sub-optimal performance whereas considering a subset, improves performance. 
 
In our work, we use cut statistics~\cite{cut_stats} to select a subset of the non-abstained dataset ($\gV$) by using the domain invariant representation \big($\gZ$\big) and the pseudo labels \big($\vy_t$\big) assigned to them. Each data sample \big($\gX_t^i, \vy^i_t$\big) where \big($ \gX_t^i \in \gV$ \big) is assigned a normalized Z score \big($\vz_i$\big) as explained below. For each patient \big($ \gX_t^i \in \gG_{patient}$\big), we first find the nearest neighbors NN\big($\gX_t^i$\big) = $ \big \{ \gX_t^l : \text{where } \big(\gX_t^l, \gX_t^i\big)$ are $K$ nearest neighbors based on $L2$ distance between $\gZ\big(\gX_t^l\big)$, $\gZ\big(\gX_t^i\big)\big \}$. A  graph \big($G =(\gV,\gE))$ is created with the number of nodes equal to the number of non-abstained patient genomic profile ($\gV$) and edges ($\gE$) defined as the nearest neighbor for each sample \big(NN\big($\gX_t^i$\big), $\gX_t^i \in \gV$ \big). For every edge in the graph a weight ($\vw_{i,j}$) is assigned, so that samples with similar representation ($\gZ$) has higher weight compared to dissimilar ones i.e., $\vw_{i,j} =  (1 + ||\gZ\big(\gX_t^i\big) - \gZ\big(\gX_t^j\big)||)^{-1}$ where $\gX_t^j \in NN\big(\gX_t^i\big)$. In general a set of data points (sub-graph) with similar representation (higher $\vw_{i,j}$) but sharing different pseudo labels are considered to be noisy and should not be considered for downstream training~\cite{cut_stats}. Under given assumption, each sample $\gX_t^i$ is assigned a score $\gJ^i$, a sum of weights of samples sharing different class labels \big($\vone\big(\vy^i_t \neq \vy^j_t\big)$\big) among the nearest neighbor. Further, under a null hypothesis of independent assignment of class labels with probability $\gP(\vy_t)$ a Z-score ($z_i$) is calculated for $\gJ^i$ using the mean ($\mu_i$) and variance ($\sigma_i$) calculated according to \citet{cut_stats}. $\gP(\vy_t)$ is approximated by the bin counts of both positive and negative classes amongst the non-abstained samples. A smaller $\vz_i$ signifies the consistency of class labels amongst the nearest neighbors and is an indicator of less noisy pseudo labels. In our work, the non-abstained patient data is sorted based on $\vz_i$, (Eq.~\ref{eqn:subset}) and the top $b\%$ (also referred to as budget) is used to obtain $\gD^{sub}_{patient}$ which is then used in conjunction with labeled cell line data to train the final classifier for drug response prediction.  
\begin{align}
    \tiny
 \vz_i &= \frac{\gJ_i - \mu_i} {\sigma_i} \nonumber  \\
 \gJ_i &= \sum\limits_{j\in \text{NN}\big(\gX_t^i\big)} \vw_{i,j} \vone\big(\vy^i_t \neq \vy^j_t\big) \nonumber \\ 
    \mu_i &= \big[1-\gP\big(\vy^i_t\big)\big]\sum\limits_{j\in \text{NN}\big(\gX_t^i\big)} \vw_{i,j} \nonumber \\
    \sigma_i^2 &= {\gP(\vy^i_t)\big[1-\gP(\vy^i_t)\big]}\sum\limits_{j\in \text{NN}\big(\gX_t^i\big)} \vw_{i,j}^2
    \label{eqn:subset}
\end{align}

Algorithm-\ref{algo:WISER} (In the appendix) describes the complete procedure of our method called the WISER (Weak supervISion and
supErvised Representation learning).

\begin{table*}[htp!]
\centering
\caption{Performance comparison of predicted patient response using AUROC and AUPRC metrics of our proposed method (\textbf{WISER}). Data related to clinical relapse is used for all the evaluations. The result is noted in the form of (mean / std) where the score has been obtained over five fold cross validation. The best performer among all baselines is reported in bold, while the predictions that were not meaningful are denoted by `-'. On an average, our method outperforms others baselines on all the drugs for at least one metric. The best performer is highlighted in \textbf{bold}.}
\resizebox{\linewidth}{!}{
\begin{tabular}{l|lrlrrlrrlrrlrr}
\hline
\textbf{Methods} & \multicolumn{2}{c}{5-Fluorouracil} &  & \multicolumn{2}{c}{Temozolomide} &  & \multicolumn{2}{c}{Sorafenib} &  & \multicolumn{2}{c}{Gemcitabine} &  & \multicolumn{2}{c}{Cisplatin} \\
\hline
 & AUROC & \multicolumn{1}{l}{AUPRC} &  & \multicolumn{1}{l}{AUROC} & \multicolumn{1}{l}{AUPRC} &  & \multicolumn{1}{l}{AUROC} & \multicolumn{1}{l}{AUPRC} &  & \multicolumn{1}{l}{AUROC} & \multicolumn{1}{l}{AUPRC} &  & \multicolumn{1}{l}{AUROC} & \multicolumn{1}{l}{AUPRC} \\
 \hline
\textbf{WISER} &{0.715/0.036} & \multicolumn{1}{l}{\textbf{0.741/0.023}} &  & \multicolumn{1}{l}{\textbf{0.760/0.006}} & \multicolumn{1}{l}{\textbf{0.786/0.019}} &  & \multicolumn{1}{l}{\textbf{0.727/0.007}} & \multicolumn{1}{l}{0.728/0.024} &  & \multicolumn{1}{l}{\textbf{0.649/0.037}} & \multicolumn{1}{l}{\textbf{0.752/0.002}} &  & \multicolumn{1}{l}{\textbf{0.851/0.007}} & \multicolumn{1}{l}{\textbf{0.861/0.020}} \\

CODE-AE & \textbf{0.868/0.030} & \multicolumn{1}{l}{0.740/0.006} &  & \multicolumn{1}{l}{0.751/0.017} & \multicolumn{1}{l}{0.762/0.001} &  & \multicolumn{1}{l}{0.631/0.020} & \multicolumn{1}{l}{0.705/0.062} &  & \multicolumn{1}{l}{0.594/0.016} & \multicolumn{1}{l}{0.751/0.006} &  & \multicolumn{1}{l}{0.652/0.071} & \multicolumn{1}{l}{0.743/0.011} \\
DAE & 0.591/0.066 & \multicolumn{1}{l}{0.573/0.066} &  & \multicolumn{1}{l}{0.685/0.013} & \multicolumn{1}{l}{0.668/0.105} &  & \multicolumn{1}{l}{0.485/0.053} & \multicolumn{1}{l}{0.613/0.046} &  & \multicolumn{1}{l}{0.530/0.036} & \multicolumn{1}{l}{0.511/0.048} &  & \multicolumn{1}{l}{0.522/0.087} & \multicolumn{1}{l}{0.581/0.096} \\
CORAL & 0.578/0.015 & \multicolumn{1}{l}{0.651/0.135} &  & \multicolumn{1}{l}{0.675/0.020} & \multicolumn{1}{l}{0.654/0.020} &  & \multicolumn{1}{l}{0.491/0.023} & \multicolumn{1}{l}{0.616/0.048} &  & \multicolumn{1}{l}{0.597/0.030} & \multicolumn{1}{l}{0.544/0.037} &  & \multicolumn{1}{l}{0.617/0.072} & \multicolumn{1}{l}{0.617/0.124} \\
VELODROME & 0.598/0.054 & 0.403/0.000 &  & 0.701/0.028 & 0.668/0.000 &  & 0.505/0.029 & \textbf{0.749/0.000} &  & 0.547/0.030 & 0.434/0.000 &  & 0.583/0.029 & 0.442/0.000 \\
ENET & 0.435/0.092 & 0.454/0.070 &  & - & - &  & - & - &  & - & - &  & 0.637/0.076 & 0.623/0.045 \\
TCRP & 0.596/0.080 & 0.546/0.073 &  & 0.675/0.009 & 0.662/0.012 &  & 0.441/0.053 & 0.521/0.054 &  & 0.462/0.057 & 0.502/0.055 &  & 0.414/0.048 & 0.432/0.037 \\
MLP & 0.569/0.050 & 0.599/0.042 &  & 0.646/0.022 & 0.624/0.038 &  & 0.444/0.035 & 0.501/0.035 &  & 0.467/0.036 & 0.498/0.049 &  & 0.459/0.070 & 0.496/0.070 \\
DSN-DANN & 0.635/0.065 & 0.596/0.101 &  & 0.683/0.015 & 0.690/0.040 &  & 0.533/0.050 & 0.628/0.069 &  & 0.555/0.070 & 0.582/0.044 &  & 0.585/0.103 & 0.608/0.133 \\
VAEN & 0.633/0.157 & 0.585/0.100 &  & 0.648/0.035 & 0.632/0.162 &  & 0.600/0.021 & 0.668/0.112 &  & 0.526/0.087 & 0.618/0.223 &  & 0.694/0.049 & 0.698/0.065 \\
COXEN & 0.336/0.000 & 0.403/0.000 &  & 0.726/0.000 & 0.668/0.000 &  & 0.639/0.000 & \textbf{0.749/0.000} &  & 0.378/0.000 & 0.434/0.000 &  & 0.393/0.000 & 0.442/0.000 \\
COXRF & 0.562/0.070 & 0.598/0.063 &  & 0.388/0.080 & 0.451/0.031 &  & 0.418/0.072 & 0.505/0.044 &  & 0.506/0.078 & 0.506/0.037 &  & 0.554/0.074 & 0.564/0.065 \\
CELLIGNER & 0.536/0.060 & 0.531/0.024 &  & - & - &  & - & - &  & 0.575/0.029 & 0.529/0.053 &  & 0.497/0.042 & 0.550/0.033 \\
ADAE & 0.68/0.040 & 0.725/0.036 &  & 0.707/0.010 & 0.757/0.003 &  & 0.540/0.092 & 0.678/0.040 &  & 0.499/0.093 & 0.691/0.123 &  & 0.633/0.165 & 0.755/0.080 \\
DSN-MMD & 0.678/0.074 & 0.674/0.103 &  & 0.712/0.031 & 0.759/0.051 &  & 0.515/0.036 & 0.582/0.090 &  & 0.465/0.041 & 0.491/0.069 &  & 0.650/0.023 & 0.605/0.067 \\
\hline
\end{tabular}

}

\label{tab:AUROC_AUPRC}
\end{table*}

\section{Experiments}

\subsection{Experiment Settings}
We evaluate the proposed method in \textbf{Four} experimental settings - (1) Drug response prediction: In this task we compare different baselines by training a binary classifier to predict efficacy of a given drug on patients, (2) understanding the medical relevance of weak supervision and subset selection techniques in this context, 
 (3) ablation study of the proposed method to compare the performance  of the model with and without weak supervision and (4) measuring sensitivity of subset size on classification performance.  

\textbf{Data} We have used the cancer cell lines and patient genomic profiles (comprising gene expression data from 1426 genes) as in CODE-AE~\cite{code-AE}. 677 labeled cancer cell line samples, from DepMap portal~\cite{deepmap}, and 9808 unsupervised patient samples from TCGA~\cite{tcga} were used. 179 samples of labeled TCGA genomic profiles were used for evaluation. Drug response in cell lines was based on z-score calculated on Area Under the Dose Response Curve (AUDRC) scores. Cell lines with a z-score less than 0 were considered positive respondents and greater than 0 as negative respondents to the drug. For patients, the assessment relied on cancer relapse time post-chemotherapy, categorizing values greater than the median as positive respondents and those less than median as negative respondents.
The specifics of data preprocessing and related details are available in~\citet{code-AE}.  
A set of 20 drugs present in both DepMap and TCGA \big($\gD = \{ d_1, d_2, \hdots d_{20}\}$\big), were considered for the experiment. Details of drugs are provided in Appendix~\ref{appx:train_detail}.  Due to the limited number of labeled patient genomic profiles, the evaluation was done only on 5-Fluorouracil (Fu), Temozolomide (Tem), Sorafenib (Sor), Gemcitabine (Gem) and Cisplatin (Cis), with drug responses available in atleast 20 patients.

\textbf{Model Configuration} The encoder and decoder networks, used in representation learning, consist of two linear layers of the neural network. The hidden units associated with the encoder and decoder are (512, 256) and (256, 512) dimensions respectively. Both networks use ReLU based activation units. The critic network and the classifier (used for weak supervision and downstream drug response prediction) consist of two layers of neural network with (64, 32) dimensions of hidden unit with ReLU activation for the first layer. The critic network uses linear layer as final activation, while the classifier uses a sigmoid layer. Same architecture has been used for all the baseline methods for fair comparison. Further details about training and hyper parameter tuning is provided in Appendix~\ref{appx:train_detail}.

\textbf{Baselines} We have compared our method with CODE-AE~\cite{code-AE}, VAEN~\cite{jia2021deep} and DAE~\cite{DAE}. Further the proposed method is compared with domain adaptation techniques like Celligner~\cite{warren2021global}, Velodrome~\cite{sharifi2021out}, Deep CORAL~\cite{CORAL} and DSN (MMD and DANN variant)~\cite{DSN}. Recent methods like ADAE~\cite{deconfounding-AE}, COXEN + Random Forest (COXRF) and COXEN~\cite{COXEN} were also included for comparison. To compare with algorithms which do not use representation learning, the results of TCRP~\cite{ma2021few}, MLP~\cite{sakellaropoulos2019deep} and ElasticNet~\cite{ElasticNet} were also included.    

\textbf{Metrics} For comparison, we have used area under the receiver operating characteristics (AUROC) and area under the precision-recall curve (AUPRC) scores~\cite{code-AE}. The classifer used for drug response prediction was trained using 5-fold stratified validation data of cell line and tested on patient data from TCGA. 
\begin{figure*}[h]
\centering
\begin{subfigure}[t]{0.21\textwidth}
    \includegraphics[keepaspectratio, width=\textwidth]{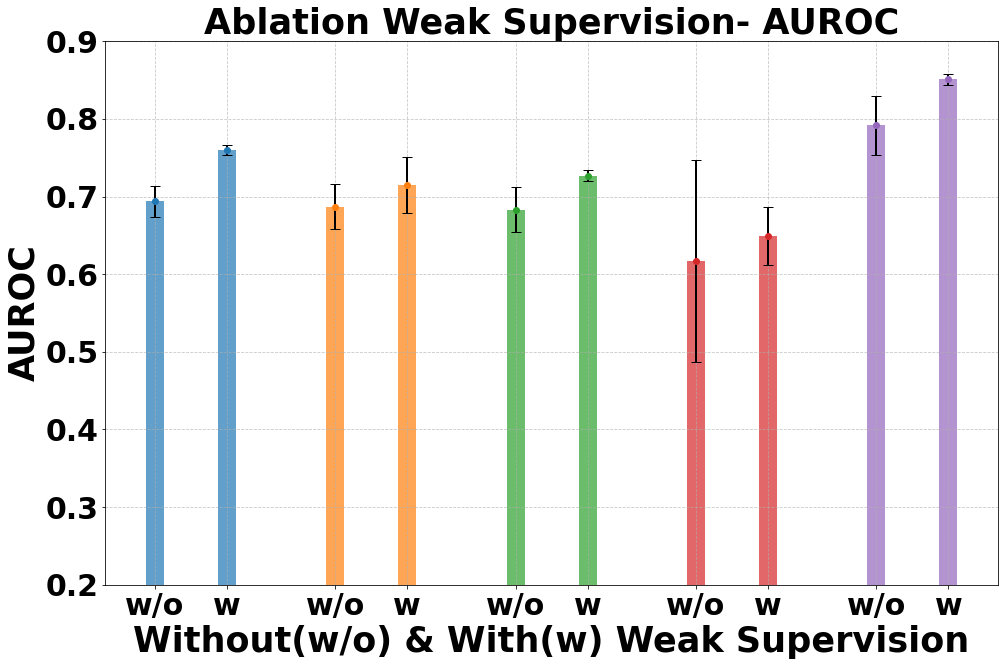}
\end{subfigure}
~
\begin{subfigure}{0.25\textwidth}
    \includegraphics[keepaspectratio, width=\textwidth]{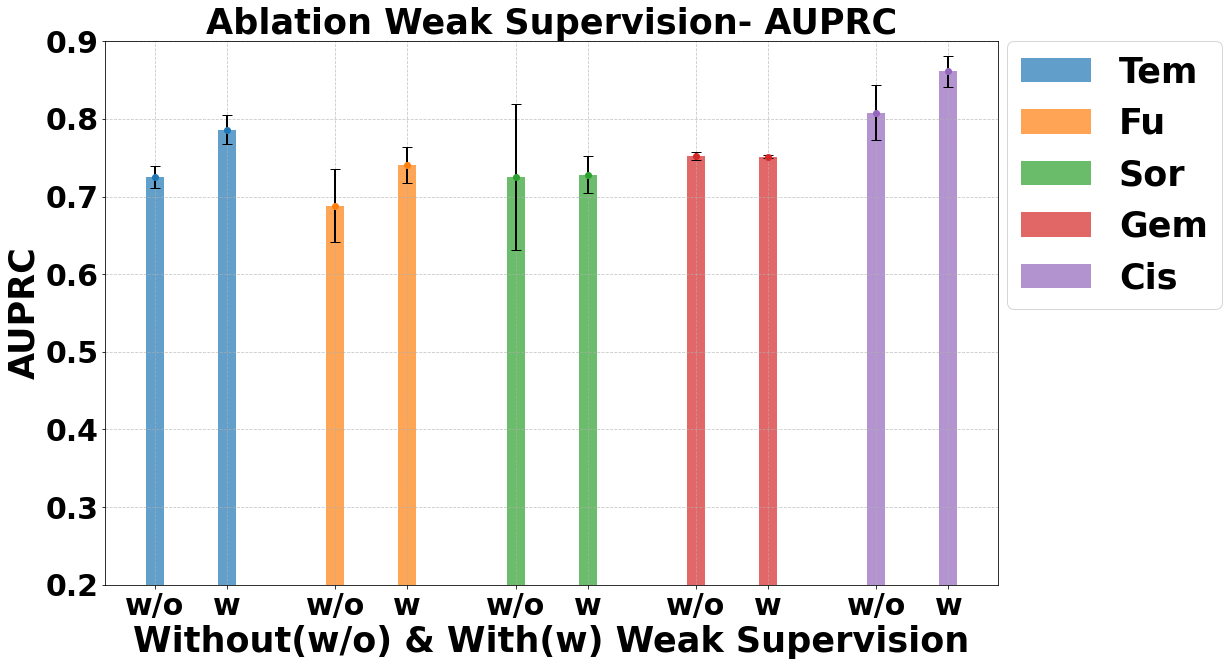}
\end{subfigure}
~
\begin{subfigure}{0.21\textwidth}
    \includegraphics[keepaspectratio, width=\textwidth]{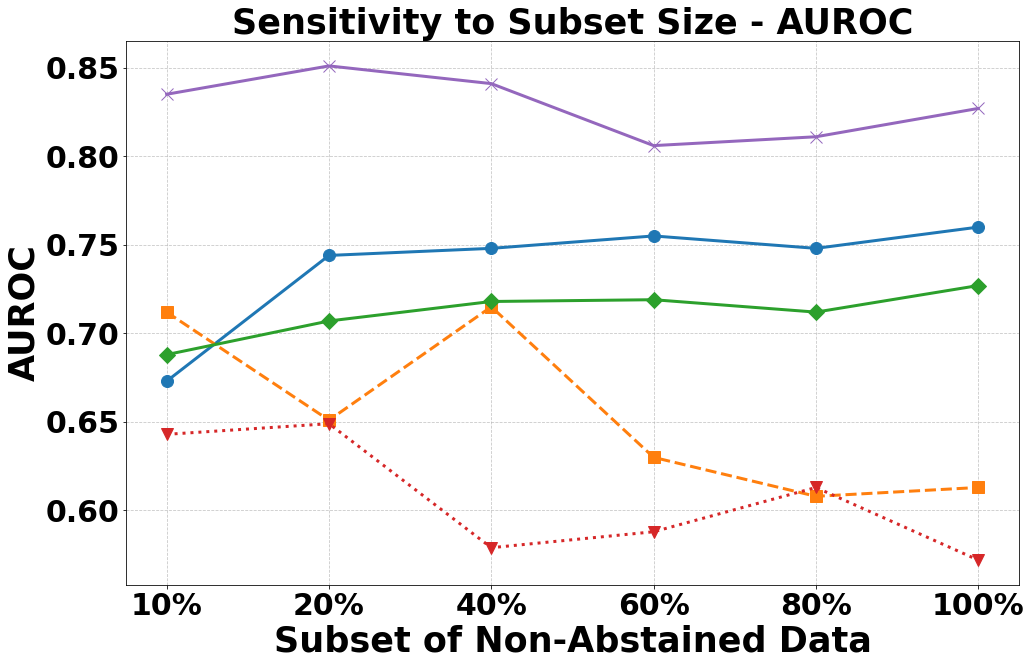}
\end{subfigure}
~
\begin{subfigure}{0.25\textwidth}
    \includegraphics[keepaspectratio, width=\textwidth]{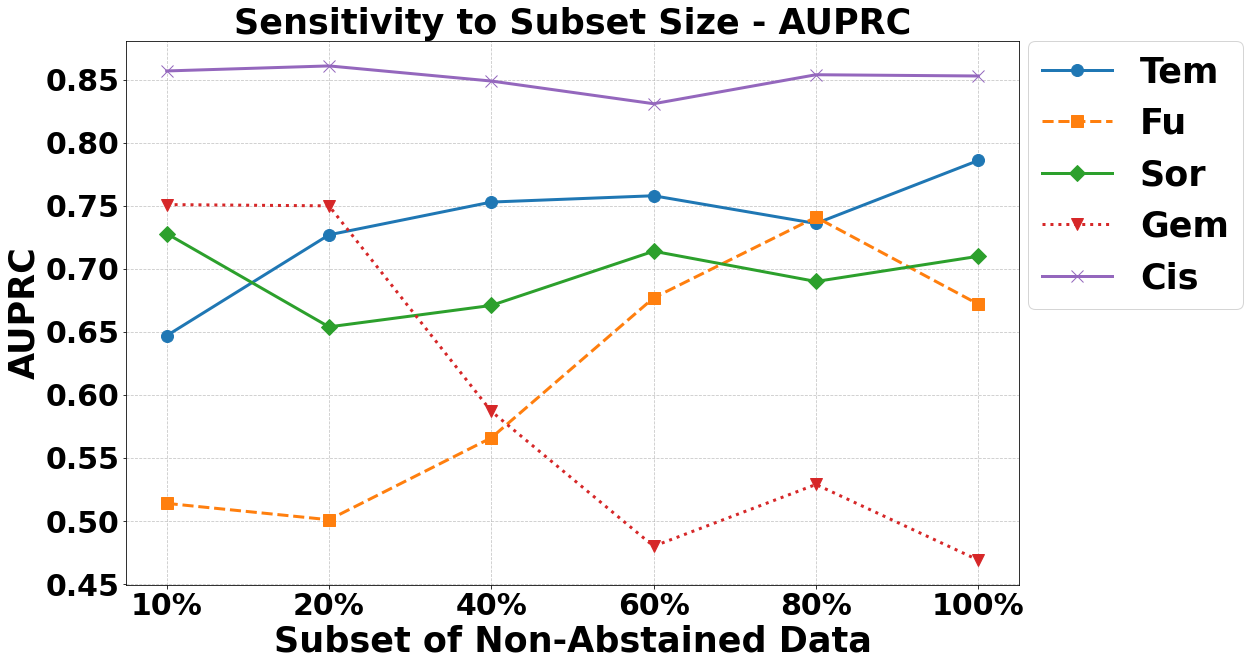}
\end{subfigure}
\caption{ Ablation on weak supervision and sensitivity test on subset size over the performance of the model. }
\label{fig:sensitivity_budget}
\vspace{-0.5cm}
\end{figure*}
\subsection{Results}
\subsubsection{Drug response prediction}
Table~\ref{tab:AUROC_AUPRC} shows a performance comparison of our method with other baselines.
Our method (WISER) exhibits superior performance in terms of AUROC scores for Cisplatin, Temozolomide, Gemcitabine, and Sorafenib, surpassing baselines by ~15.7\%, 0.9\%, 5.2\% and 8.8\% respectively while for AUPRC score it shows an enhancement of 0.1\%, 2.4\%, 0.1\% and 10.6\% for 5-Fluorouracil, Temozolomide, Gemcitabine and Cisplatin respectively. Comparison with other traditional methods is provided in Appendix~\ref{appx:tradional_methods}.

\subsubsection{Medical Relevance of the Method}
We next examine the medical relevance of the pseudo labels and the subset selected for the downstream prediction, generated using the best hyperparameters. We identify the genes most relevant in the generation of these pseudo labels, through the feature selection procedure of Extra-Trees Classifier~\cite{alfian2022predicting}. This is done by fitting an Extra-Trees Classifier model on the patient genomic profiles (from the selected subset) and their pseudo labels, and selecting genes with top 50\% feature importance. We compare the selected genes against the GDISC database~\cite{spainhour2017gdisc}, which has independently identified genes associated with chemotherapy response in TCGA. The authors provide information for 22 drugs, however, details for Sor were unavailable.
The resulting set of significant genes and the corresponding overlap are highlighted in Table~\ref{tab:med_relevance}. Drugs with relevant information were evaluated on two metrics (1) \textbf{Precision}:  
 This measures the ratio of genes marked as significant by GDISC among all the genes selected by the Extra-Trees Classifier. (2) \textbf{Recall}: This assesses the ratio of genes selected by the Extra-Trees Classifier among the entire set of genes marked as significant by the GDISC database for a given drug. Cisplatin, Temozolomide, Gemcitabine, 5-Fluorouracil
achieve a precision of 0.860, 0.609, 0.499, 0.419 respectively. Similarly, the recall achieved  by these drugs are 0.503, 0.500, 0.464, 0.459. This score correlates with the performance of the drugs in Table~\ref{tab:AUROC_AUPRC}, where a higher precision and recall generate better AUROC and AUPRC, thus suggesting the faithfulness \cite{alvarez2018towards} of our explanations in terms of gene importances.  
\subsubsection {Ablation Studies}
\begin{table}
\caption{ Experiment to examine the medical relevance of weak supervision and subset selection. In the given experiment, the set of genes with significant drug-gene interaction (P-val$<$0.05) for the survival of patients with cancer, from GDISC, is compared with the genes considered relevant by weak supervision and subset selection. The precision and recall between the two sets is reported.}
\resizebox{\linewidth}{!}{
\begin{tabular}{c|ccccc}
\hline
Drug &Fu & Tem & Sor & Gem & Cis\\
\hline
Gene (P-val$<$0.05) & 418   & 706   & - & 526   & 831 \\  

Gene (P-val$\geq$0.05) & 521   & 456   & - & 473   & 143   \\
\hline
Precision         & 0.419 & 0.609 & - & 0.499 & 0.860  \\
Recall            & 0.459 & 0.500   & - & 0.464 & 0.503 \\
   \hline
\end{tabular}
}
\label{tab:med_relevance}
\vspace{-0.86cm}
\end{table}
We conducted an ablation test on the effect of weak supervision and subset selection, by directly using the representations of labeled cell line samples for the downstream drug response prediction. 
The results (Figure~\ref{fig:sensitivity_budget}) were compared for the best hyperparameter configuration of each drug. The results indicate that weak supervision and subset selection (WISER) improve AUROC by 4.58\% and AUPRC by 3.4\% on average. Further details on the experiments are provided in Appendix~\ref{sec:without_ws}. 

\subsubsection{Sensitivity Analysis}
Since the ablation study indicates the importance of weak supervision and subset selection, we next examine the impact of the subset budget (b) on the overall performance. This test was performed by varying b while maintaining the optimal configuration for the remaining parameters. Figure~\ref{fig:sensitivity_budget} summarizes the result of the experiment. For AUROC, the subset selection setting generated better results for 5-Fluorouracil (b=40\%), Cisplatin (b=20\%) and Gemcitabine (b=20\%) than the complete non-abstained dataset (b=100\%). An improvement of 10.2\%, 2.4\% and 7.7\% are seen in these 3 drugs respectively. For AUPRC, the subset selection setting generated better results for all drugs other than Temozolomide, with budget b set to 20\%, 10\%, 10\%, 80\% for Cisplatin, Gemcitabine, Sorafenib and 5-Fluorouracil respectively. An improvement of 0.8\%, 28.2\%, 1.8\% and 6.9\% was observed for these drugs respectively. It can be seen that using subset selection leads to optimal performance compared to the complete non-abstained dataset.

\section{Conclusion }
Recent cancer drug response prediction methods have largely followed the paradigm of unsupervised domain-invariant representation learning followed by a downstream drug response classification step. Although supervised training could improve performance, doing so was limited by the heterogeneity in patient responses across drugs and limited availability of labeled patient data. Our approach, addresses these challenges by modeling genomic profiles as a combination of discrete drug representations, reflective of heterogeneous drug responses.
We also use weak supervision and subset selection on unlabeled patient genomic profiles to improve generalization of the classifier. WISER demonstrates improved drug response prediction for several clinically significant anti-cancer drugs. To the best of our knowledge, our method is the first to use domain-invariant representation for subset selection with weak supervision, and can be applied to similar settings with large unlabeled datasets. However, the performance of our method is limited by the available labeled dataset and the set of drugs considered for discrete representation learning. Future work can explore further improvements of our approach through other sources of distant supervision, e.g., through knowledge graphs.

\section{Impact Statement}
This research seeks to enhance the effectiveness of personalized cancer treatment by integrating laboratory data and patient information, thereby bridging gaps between research and real-world outcomes. The study tackles the scarcity of labeled patient data through the use of weak supervision techniques, aiming to contribute to the improvement of reliable and accessible personalized cancer treatments.
\nocite{langley00}

\bibliography{example_paper}
\bibliographystyle{icml2024}

\newpage
\appendix
\onecolumn
\section{Appendix}

\begin{algorithm}
\scriptsize
\caption{WISER: Weak supervISion and supErvised Representation learning to improve drug response prediction in cancer}
\label{algo:WISER}
\begin{algorithmic}[1]
\REQUIRE Genomic profile for cell line ($\gX_c$), genomic profile for patients ($\gX_t$), epoch for initial training ($\gP_i$), epoch for domain adaptation based training ($\gP_d$), epoch for critic training ($\gP_c$), batch size ($\gB$), weak supervision thresholds $t^{+}$ and $t^{-}$, number of chunks $\gO$ of cell line data for training, subset selection budget $b$ and nearest neighbor size $K$.
\STATE \textbf{\#\#Representation Learning}
\FOR{epoch in [0 $\hdots \gP_i$] :}
    \STATE Sample batch of cell line and patient genomic data from the dataloader without replacement. $\{ \gX_c^{(i)} \}_{i=0} ^{\gB}$ $\{ \gX_t^{(j)} \}_{j=0} ^{\gB}$
    \STATE Train shared encoder ($\gC_{\gS}$), private encoder ($\gC_{\gP}^c$, $\gC_{\gP}^t$), discrete embedding ($\gR$), and decoder ($\gD$) with sampled batch using ($\vl_{pl}$) loss. 
\ENDFOR
\STATE $\gN$ = 0
\FOR{epoch in [0 $\hdots \gP_d$] :}
    \STATE Sample batch of cell line and patient genomic data from the dataloader without replacement. $\{ \gX_c^{(i)} \}_{i=0} ^{\gB}$ $\{ \gX_t^{(j)} \}_{j=0} ^{\gB}$
    \STATE Train the critic network ($\gF$) with ($\vl_{critic}$) loss. 
    \STATE $\gN$+=1
    \IF{$ \gN\% \gP_c == 0$}
        \STATE Sample batch of genomic data $\{ \gX_c^{(i)} \}_{i=0} ^{\gB}$ $\{ \gX_t^{(j)} \}_{j=0} ^{\gB}$ 
        \STATE Train $\gC_{\gS}, \gC_{\gP}^c, \gC_{\gP}^t, \gR, \gD $ with sampled batch using $\vl_{total}$ )
    \ENDIF
\ENDFOR
\STATE Use the representation ($\gZ$) generated by shared encoder and drug-based embeddings ($\gC_{\gS} \text{ and } \gR$).
\STATE \textbf{\#\# Weak Supervision}
\FOR{i in [1 $\hdots$ $\gO$]:}
    \STATE Train a classifier $\gM_{i}$ using $D^{i}_{cell}$ where $D^{i}_{cell} \subset G_{cell}$.
    \STATE Infer $\gP_{i}(\vy|\gX_{t}^{j})$ using the trained classifier $\gM_{i}$, where $\gX_{t}^{j} \in \gG_{patient}$.
\ENDFOR
\STATE Label samples based on $t^{+}$ and $t^{-}$ (Eq. \ref{eqn:WS}).
\STATE Assign the final pseudo label($\vy^{j}_t$), based on Majority Voting strategy, for non-abstained samples (Eq. \ref{eqn:MV}).
\STATE \textbf{\#\# Subset Selection and Drug Response Prediction}
\STATE Calculate $\vz_{i}$ for non-abstained patient samples as in Eq. \ref{eqn:subset}, sort by $\vz_{i}$ and choose top $b$ \% as the subset.
\STATE Use the patient genomic profiles associated with this subset, alongwith their pseudo labels, in conjunction with $\gX_{c}$ to train a drug response prediction classifier.
\end{algorithmic}
\end{algorithm}

\section{Distinction between the two domains}
\label{appx: domain_details}
Table~\ref{tab:domain_details} provides details of the two domains under consideration in our study. The cell line domain is notable for its abundant labeled responses to diverse drugs, whereas the patient data predominantly comprises unlabeled samples. For our experiments, we selected 20 drugs which were administered in both patients and cell lines. To evaluate our approach on patients, we considered 5 drugs with a documented response in at least 20 patients (Table \ref{tab:test_samples}).

\begin{table}[hp!]
\caption{Details about the two domains in cancer drug response prediction.}
\resizebox{\linewidth}{!}{
\begin{tabular}{p{1.2cm}|p{1.2cm}p{1.2cm}p{4cm}p{3cm}p{3.5cm}}
\hline
Domains  & Unlabeled data   & Labeled data                    & Drug response label  & Number of drugs with response  & Number of drugs selected in our experiments \\
\hline
Cell line    & 1305 & 686 & Evaluated using Z-score computed on AUDRC scores. (1) Z-score less than 0 considered as positive respondents. (2) Z-score greater than 0 considered as negative respondents.  &  449 & 20                       \\
\hline
Patients & 9808  & 179 & Cancer relapse time post-chemotherapy (1) Values greater than the median considered positive respondents. (2) Values less than the median considered negative respondents. & 78 & 5 \\
\hline
\end{tabular}}
\label{tab:domain_details}
\end{table}

\begin{table}[]
\caption{Distribution of testing dataset.}
\centering
\begin{tabular}{@{}llllll@{}}
\hline
\textbf{Drug}         & 5-Fluorouracil & Temozolomide & Gemcitabine & Cisplatin & Sorafenib \\ 
\hline
\textbf{TCGA samples} & 21             & 46           & 46          & 40        & 26        \\ 
\hline
\end{tabular}
\label{tab:test_samples}
\end{table}

\section{Training details and hyperparameter}
\label{appx:train_detail}
The training of the model happens in four stages as mentioned in Algorithm~\ref{algo:WISER}. For representation learning, a grid search was performed on the initial training epoch \big($\gP_i$\big), domain adaptation epoch\big($\gP_d$\big) and inverse temperature value ($\Delta$). The value considered for the experiments were \big[50, 100, 300\big], \big[1000, 2000, 2500, 3000\big] and \big[0.001, 0.1, 1, 2, 2.5, 10, 100\big] respectively. A set of 20 drugs were used for representation learning ($\gR$) namely 5-Fluorouracil, Gemcitabine, Temozolomide, Cisplatin, Sorafenib, Sunitinib, Doxorubicin, Tamoxifen, Paclitaxel, Carmustine, Cetuximab, Methotrexate, Topotecan, Erlotinib, Irinotecan, Bicalutamide, Temsirolimus, Oxaliplatin, Docetaxel, Etoposide. For weak supervision, 5 label functions were trained on 5 different chunks ($\gO$) of labeled cell line dataset. The number of chunk for training was decided based on previous works~\cite{code-AE,ratner2017snorkel}, considering the limited number of labeled cell line data and optimal performance of majority vote for less than 10 (Lfs). The value of ($t^+$ and $t^-$) were determined based on grid search over  \big[(0.7, 0.3), (0.55, 0.49), (0.51, 0.46)\big] respectively. Similar experiments were performed on the median score of the predicted probabilities of label functions. 
For subset selection, K=20 was considered in line with previous work~\cite{subset_ws}. The optimal value of subset size (b) was determined by a grid search over \big[0.2, 0.4, 0.5, 0.6, 0.8, 1\big]. All the experiments were done on  NVIDIA A6000 Graphic card with 20 core and 160 GB memory.
Our code is available at  \href{https://github.com/kyrs/WISER}{https://github.com/kyrs/WISER}. 
\section{Experiment without weak supervision}
\label{sec:without_ws}
\begin{table*}[htp!]
\centering
\caption{Performance comparison of predicted patient response using AUROC and AUPRC metrics of our proposed method without weak supervision (\textbf{WISER(w\textbackslash o WS)}) with other transfer learning based approaches. Data related to clinical relapse is used for all the evaluations. The result is noted in the form of (mean / std) where the score has been obtained over five fold cross validation. The best performer among all baselines is reported in bold, while the predictions that were not meaningful are denoted by `-'. On an average, our method performs the best on 3 out of 5 drugs for atleast one metric. The best performer is highlighted in \textbf{bold}.}
\resizebox{\linewidth}{!}{
\begin{tabular}{l|lrlrrlrrlrrlrr}
\hline
\textbf{Methods} & \multicolumn{2}{c}{5-Fluorouracil} &  & \multicolumn{2}{c}{Temozolomide} &  & \multicolumn{2}{c}{Sorafenib} &  & \multicolumn{2}{c}{Gemcitabine} &  & \multicolumn{2}{c}{Cisplatin} \\
\hline
 & AUROC & \multicolumn{1}{l}{AUPRC} &  & \multicolumn{1}{l}{AUROC} & \multicolumn{1}{l}{AUPRC} &  & \multicolumn{1}{l}{AUROC} & \multicolumn{1}{l}{AUPRC} &  & \multicolumn{1}{l}{AUROC} & \multicolumn{1}{l}{AUPRC} &  & \multicolumn{1}{l}{AUROC} & \multicolumn{1}{l}{AUPRC} \\
 \hline
\textbf{WISER- w\textbackslash o WS} &{0.687/0.029} & \multicolumn{1}{l} {0.688/0.047} &  & \multicolumn{1}{l}{0.694/0.020} & \multicolumn{1}{l}{0.725/0.014} &  & \multicolumn{1}{l}{\textbf{0.683/0.021}} & \multicolumn{1}{l}{0.725/0.094} &  & \multicolumn{1}{l}{\textbf{0.617/0.013}} & \multicolumn{1}{l}{\textbf{0.752/0.005}} &  & \multicolumn{1}{l}{\textbf{0.792/0.038}} & \multicolumn{1}{l}{\textbf{0.808/0.035}} \\

CODE-AE & \textbf{0.868/0.030} & \multicolumn{1}{l}{\textbf{0.740/0.006}} &  & \multicolumn{1}{l}{\textbf{0.751/0.017}} & \multicolumn{1}{l}{\textbf{0.762/0.001}} &  & \multicolumn{1}{l}{0.631/0.020} & \multicolumn{1}{l}{0.705/0.062} &  & \multicolumn{1}{l}{0.594/0.016} & \multicolumn{1}{l}{0.751/0.006} &  & \multicolumn{1}{l}{0.652/0.071} & \multicolumn{1}{l}{0.743/0.011} \\
TCRP & 0.596/0.080 & 0.546/0.073 &  & 0.675/0.009 & 0.662/0.012 &  & 0.441/0.053 & 0.521/0.054 &  & 0.462/0.057 & 0.502/0.055 &  & 0.414/0.048 & 0.432/0.037 \\
CORAL & 0.578/0.015 & \multicolumn{1}{l}{0.651/0.135} &  & \multicolumn{1}{l}{0.675/0.020} & \multicolumn{1}{l}{0.654/0.020} &  & \multicolumn{1}{l}{0.491/0.023} & \multicolumn{1}{l}{0.616/0.048} &  & \multicolumn{1}{l}{0.597/0.030} & \multicolumn{1}{l}{0.544/0.037} &  & \multicolumn{1}{l}{0.617/0.072} & \multicolumn{1}{l}{0.617/0.124} \\
VELODROME & 0.598/0.054 & 0.403/0.000 &  & 0.701/0.028 & 0.668/0.000 &  & 0.505/0.029 & \textbf{0.749/0.000} &  & 0.547/0.030 & 0.434/0.000 &  & 0.583/0.029 & 0.442/0.000 \\
\hline
\end{tabular}

}
\label{tab:AUROC_AUPRC_wo_ws}
\end{table*}

To assess the impact of incorporating labeled drug response data for representation learning, we conducted a separate experiment with our method, excluding weak supervision and subset selection. The results were compared with other transfer learning-based approaches, and the findings are presented in Table \ref{tab:AUROC_AUPRC_wo_ws}. Our method (\big(WISER w\textbackslash o WS \big)) demonstrated superior performance in three out of five drugs, showcasing improved AUROC and AUPRC metrics. Specifically, it exhibited gains of approximately 14\%, 2.3\%, and 5.2\% in AUROC for Cisplatin, Gemcitabine, and Sorafenib, respectively. Additionally, there were gains of 6.5\% and 0.1\% in AUPRC for Cisplatin and Gemcitabine, respectively.

\subsection{Sensitivity test on hyperparameter}
To analyze the impact of different hyperparameters on representation learning, we have done a sensitivity analysis of initial training epoch ($\gP_i$), adversarial training ($\gP_d$) and temperature ($\Delta$) on AUROC and AUPRC performance while using best configuration for other hyper parameters .

\subsubsection{Initial training epoch}
Table~\ref{tab:sen_initial_epoch_AUROC} and Table~~\ref{tab:sen_initial_epoch_AUPRC} shows the impact of initial training epoch on the performance of the model. As per the result the Cisplatin, Gemcitabine, Temozolomide, Sorafenib, and 5-Fluorouracil achieves best AUROC score for 50, 300, 300, 50, 50 epoch respectively and best AUPRC score for 50, 100, 300, 100, 100 respectively. In general, different drugs performs differently for this hyper parameter, where training for less number of epochs is favourable for Cisplatin  while training for more iteration is favoured in Temozolomide.
\begin{table*}[h]
\centering
\caption{Sensitivity analysis of the initial training epoch \big($\gP_i$\big) on AUROC scores.}
\begin{tabular}{c|ccc} 
\hline
Drug  /\ Epoch & 50 & 100 & 300 \\
\hline
Fu  & \textbf{0.687/0.027} & 0.642/0.043 & 0.650/0.034 \\
Sor & \textbf{0.683/0.019} & 0.636/0.047 & 0.441/0.043 \\
Tem & 0.558/0.004 & 0.573/0.024 & \textbf{0.694/0.017} \\
Gem & 0.499/0.097 & 0.440/0.120 & \textbf{0.617/0.011} \\
Cis & \textbf{0.792/0.034} & 0.527/0.045 & 0.529/0.120 \\
\hline
\end{tabular}
\label{tab:sen_initial_epoch_AUROC}
\end{table*}

\begin{table*}[htp!]
\centering
\caption{Sensitivity analysis of the initial training epoch \big($\gP_i$\big) on AUPRC scores.}
\begin{tabular}{c|ccc}
 \hline
Drug /\ Epoch & 50 & 100 & 300 \\
\hline
Fu  & 0.626/0.069 & \textbf{0.688/0.042} & 0.575/0.022 \\
Sor & 0.431/0.100 & \textbf{0.725/0.085} & 0.566/0.065 \\
Tem & 0.547/0.029 & 0.575/0.024 & \textbf{0.725/0.014} \\
Gem & 0.75/0.0    & \textbf{0.752/0.005} & 0.75/0.0    \\
Cis & \textbf{0.808/0.031} & 0.575/0.057 & 0.608/0.081\\
\hline
\end{tabular}
\label{tab:sen_initial_epoch_AUPRC}
\end{table*}

\subsubsection{Adversarial training Epoch}
Next we analyze the impact of adversarial training epoch on the performance of the model. Table~\ref{tab:sen_adv_epoch_AUROC} and Table~\ref{tab:sen_adv_epoch_AUPRC} shows the result of given experiment. Where, Cisplatin, Gemcitabine, Temozolomide, Sorafenib, and 5-Fluorouracil achieves best AUROC score for 2500, 2000, 1000, 2000, 2500 respectively and best AUPRC score for 2000, 1000,1000, 2000, 2000 respectively. In general, we see an impact of domain adaptation on the performance of the model, where in drugs like Cisplatin best results are generated for larger number of training epochs.

\begin{table}[H]
\centering
\caption{Sensitivity analysis of the domain adversarial training epoch ($\gP_d$) on AUROC scores.}
\begin{tabular}{c|ccc}
\hline
Drug /\ Epoch &1000 & 2000 & 2500 \\
\hline
Fu  & 0.509/0.063 & 0.680/0.013 & \textbf{0.687/0.026} \\
Sor & 0.452/0.061 & \textbf{0.683/0.019} & 0.586/0.027 \\
Tem & \textbf{0.694/0.018} & 0.681/0.010 & 0.664/0.031 \\
Gem & 0.518/0.082 & \textbf{0.617/0.011} & 0.517/0.111 \\
Cis & 0.391/0.011 & 0.654/0.085 & \textbf{0.792/0.033} \\
\hline
\end{tabular}
\label{tab:sen_adv_epoch_AUROC}
\end{table}

\begin{table}[H]
\centering
\caption{Sensitivity analysis of the domain adversarial training epoch ($\gP_d$) on AUPRC scores.}
\begin{tabular}{c|ccc}
 \hline
Drug /\ Epoch & 1000 & 2000 & 2500 \\
\hline
Fu  & 0.589/0.108 & \textbf{0.688/0.042} & 0.656/0.0348 \\
Sor & 0.523/0.113 & \textbf{0.725/0.085} & 0.565/0.101  \\
Tem & \textbf{0.725/0.013} & 0.683/0.035 & 0.659/0.048  \\
Gem & \textbf{0.750/0.000}    & -           &\textbf{0.750/0.000}     \\
Cis & 0.525/0.016 & 0.710/0.058 & \textbf{0.808/0.031} \\ 
\hline
\end{tabular}
\label{tab:sen_adv_epoch_AUPRC}
\end{table}
\subsubsection{Inverse temperature}
We have further conducted a sensitivity test to analyze the  the inverse temperature ($\Delta$).  Table~\ref{tab:sen_temp_AUROC}  and Table~\ref{tab:sen_temp_AUPRC} summarizes the result of the given
experiment. Based on the presented outcome, we found an influence of inverse temperature on the performance of all the
drugs. Inverse temperature controls the weights associated with the drug embeddings ($\gR$).  For 5-Fluorouracil and Temozolomide best AUROC and AUPRC scores are generated for $\Delta = 10$, Cisplatin
 generates best results for $\Delta = 2.5$ while for Gemcitabine, Sorafenib best AUPRC score  were generated for $\Delta = 0.001$ and AUROC score for  0.1 and 2.5 respectively.  
\begin{table}[H]
\centering
\caption{Sensitivity analysis of the Inverse Temperature ($\Delta$) on AUROC scores.}
\begin{tabular}{c|cccccc}
\hline
Drug /\ Inv temp &
 0.01 &
  0.1 &
 1&
  2 &
  2.5 &
  10 \\
  \hline
Fu  & 0.487/0.094 & 0.483/0.231                         & 0.500/0.079 & 0.486/0.031 & 0.478/0.091                         &\textbf{0.687/0.026} \\
Tem & 0.492/0.07  & 0.534/0.103                         & 0.366/0.016 & 0.445/0.035 & 0.489/0.071                         & \textbf{0.694/0.017} \\
Gem & 0.465/0.035 & \textbf{0.617/0.011} & 0.390/0.015 & 0.324/0.02  & 0.330/0.014                         & 0.365/0.016                         \\
Sor & 0.483/0.141 & 0.456/0.127                         & 0.473/0.045 & 0.638/0.06  & \textbf{0.638/0.051} & 0.586/0.027                         \\
Cis & 0.526/0.057 & 0.558/0.120                         & 0.541/0.047 & 0.502/0.043 &\textbf{0.791/0.033} & 0.485/0.15   \\
\hline
\end{tabular}

\label{tab:sen_temp_AUROC}
\end{table}
\begin{table}[H]
\centering
\caption{Sensitivity analysis of the Inverse Temperature ($\Delta$) on AUPRC scores.}
\begin{tabular}{c|cccccc}
\hline
Drug /\ Inv temp & 0.01 & 0.1 & 1 & 2 & 2.5 & 10 \\
\hline
Fu  & 0.601/0.09                          & 0.542/0.177 & 0.469/0.060 & 0.499/0.057 & 0.440/0.0491                        & \textbf{0.688/0.042} \\
Tem & 0.561/0.079                         & 0.533/0.098 & 0.416/0.017 & 0.490/0.019 & 0.486/0.066                         & \textbf{0.725/0.013} \\
Gem & \textbf{0.670/0.086} & 0.481/0.072 & 0.572/0.078 & 0.424/0.027 & 0.402/0.115                         & 0.431/0.01                          \\
Sor & \textbf{0.724/0.084} & 0.609/0.162 & 0.479/0.024 & 0.553/0.083 & 0.526/0.042                         & 0.554/0.021                         \\
Cis & 0.603/0.054                         & 0.598/0.120 & 0.664/0.034 & 0.626/0.047 & \textbf{0.808/0.031} & 0.501/0.021  \\                       \hline
\end{tabular}
\label{tab:sen_temp_AUPRC}
\end{table}

\section{Analysis of different components}
\label{appx:tradional_methods}
We have further analyzed the importance of different components in our method. For this, we have compared WISER with (1) WISER- w\textbackslash o WS : A derivative of our work with only supervised discrete representation learning module and does not use weak supervision. (2) Code-AE : Code-AE is the closest baseline to our work which does not use weak supervision and  supervised discrete representation learning.  (3) Next we compare our method with other representation learning based methods without domain adaptation module i.e., Variational autoencoder~\cite{VAE} (VAE), autoencoder~\cite{hinton1993autoencoders} (AE). (4) Finally we compare our method with random forest a standard model without neural network (RF). As per the result WISER performs optimal for all the drugs on atleast one metric.

\begin{table*}[htp!]
\centering
\caption{Performance comparison of predicted patient response using AUROC and AUPRC metrics of our proposed method (\textbf{WISER}). Data related to clinical relapse is used for all the evaluations. The result is noted in the form of (mean / std) where the score has been obtained over five fold cross validation. On an average, our method outperforms others baselines on all the drugs for at least one metric. The best performer is highlighted in \textbf{bold}.}
\resizebox{\linewidth}{!}{
\begin{tabular}{l|lrlrrlrrlrrlrr}
\hline
\textbf{Methods} & \multicolumn{2}{c}{5-Fluorouracil} &  & \multicolumn{2}{c}{Temozolomide} &  & \multicolumn{2}{c}{Sorafenib} &  & \multicolumn{2}{c}{Gemcitabine} &  & \multicolumn{2}{c}{Cisplatin} \\
\hline
 & AUROC & \multicolumn{1}{l}{AUPRC} &  & \multicolumn{1}{l}{AUROC} & \multicolumn{1}{l}{AUPRC} &  & \multicolumn{1}{l}{AUROC} & \multicolumn{1}{l}{AUPRC} &  & \multicolumn{1}{l}{AUROC} & \multicolumn{1}{l}{AUPRC} &  & \multicolumn{1}{l}{AUROC} & \multicolumn{1}{l}{AUPRC} \\
 \hline
\textbf{WISER} &{0.715/0.036} & \multicolumn{1}{l}{\textbf{0.741/0.023}} &  & \multicolumn{1}{l}{\textbf{0.760/0.006}} & \multicolumn{1}{l}{\textbf{0.786/0.019}} &  & \multicolumn{1}{l}{\textbf{0.727/0.007}} & \multicolumn{1}{l}{\textbf{0.728/0.024}} &  & \multicolumn{1}{l}{\textbf{0.649/0.037}} & \multicolumn{1}{l}{\textbf{0.752/0.002}} &  & \multicolumn{1}{l}{\textbf{0.851/0.007}} & \multicolumn{1}{l}{\textbf{0.861/0.020}} \\
WISER- w\textbackslash o WS &{0.687/0.029} & \multicolumn{1}{l} {0.688/0.047} &  & \multicolumn{1}{l}{0.694/0.020} & \multicolumn{1}{l}{0.725/0.014} &  & \multicolumn{1}{l}{0.683/0.021} & \multicolumn{1}{l}{0.725/0.094} &  & \multicolumn{1}{l}{0.617/0.013} & \multicolumn{1}{l}{0.752/0.005} &  & \multicolumn{1}{l}{0.792/0.038} & \multicolumn{1}{l}{0.808/0.035} \\
CODE-AE & \textbf{0.868/0.030} & \multicolumn{1}{l}{0.740/0.006} &  & \multicolumn{1}{l}{0.751/0.017} & \multicolumn{1}{l}{0.762/0.001} &  & \multicolumn{1}{l}{0.631/0.020} & \multicolumn{1}{l}{0.705/0.062} &  & \multicolumn{1}{l}{0.594/0.016} & \multicolumn{1}{l}{0.751/0.006} &  & \multicolumn{1}{l}{0.652/0.071} & \multicolumn{1}{l}{0.743/0.011} \\
VAE & 0.636/0.032 & \multicolumn{1}{l}{0.616/0.067} &  & \multicolumn{1}{l}{0.671/0.023} & \multicolumn{1}{l}{0.688/0.020} &  & \multicolumn{1}{l}{0.472/0.023} & \multicolumn{1}{l}{0.554/0.024} &  & \multicolumn{1}{l}{0.514/0.090} & \multicolumn{1}{l}{0.484/0.048} &  & \multicolumn{1}{l}{0.552/0.103} & \multicolumn{1}{l}{0.631/0.034} \\
AE & 0.636/0.019 & 0.576/0.046 &  & 0.659/0.048 & 0.610/0.030 &  & 0.528/0.061 & 0.597/0.101 &  & 0.553/0.029 & 0.553/0.050 &  & 0.623/0.042 & 0.607/0.067 \\
RF & 0.565/0.100 & \multicolumn{1}{l}{0.595/0.099} &  & \multicolumn{1}{l}{0.632/0.03} & \multicolumn{1}{l}{0.619/0.048} &  & \multicolumn{1}{l}{0.366/0.131} & \multicolumn{1}{l}{0.482/0.103} &  & \multicolumn{1}{l}{0.452/0.026} & \multicolumn{1}{l}{0.480/0.023} &  & \multicolumn{1}{l}{0.470/0.062} & \multicolumn{1}{l}{0.473/0.044} \\
\hline
\end{tabular}

}
\label{tab:AUROC_AUPRC_sup}
\end{table*}




\section{Ablation for discrete representation}
\label{appx:loss_ablation}
We have extended our analysis by performing ablation studies to assess the significance of the proposed loss functions used in training of domain invariant representation(Z). For this experiment we successively removed triplet loss $\vl_{cns}$ and 
 and discrete representation loss $\vl_{embed}$ 
. The experiments were performed without weak supervision and subset selection strategy in the downstream drug response prediction task, so as to understand the effects of the loss terms on Z in isolation (indicated by 'Wiser(Z)'). The results (mean/std. over 5-fold cross validation) of this experiment are provided in Table \ref{tab:ablation_auroc_auprc}.

\begin{table*}[h!]
\centering
\caption{An ablation study to investigate the impact of different loss functions on AUROC and AUPRC metrics for discrete embedding. Data related to clinical relapse is used for all the evaluations. The result is noted in the form of (mean / std) where the score has been obtained over five fold cross validation. On an average, our method outperforms others baselines on all the drugs for at least one metric. The best performer is highlighted in \textbf{bold}.}
\resizebox{\linewidth}{!}{
\begin{tabular}{l|lrlrrlrrlrrlrr}
\hline
\textbf{Methods} & \multicolumn{2}{c}{5-Fluorouracil} &  & \multicolumn{2}{c}{Temozolomide} &  & \multicolumn{2}{c}{Sorafenib} &  & \multicolumn{2}{c}{Gemcitabine} &  & \multicolumn{2}{c}{Cisplatin} \\
\hline
 & AUROC & \multicolumn{1}{l}{AUPRC} &  & \multicolumn{1}{l}{AUROC} & \multicolumn{1}{l}{AUPRC} &  & \multicolumn{1}{l}{AUROC} & \multicolumn{1}{l}{AUPRC} &  & \multicolumn{1}{l}{AUROC} & \multicolumn{1}{l}{AUPRC} &  & \multicolumn{1}{l}{AUROC} & \multicolumn{1}{l}{AUPRC} \\
 \hline
\textbf{WISER(Z)} &{\textbf{0.687/0.029}} & \multicolumn{1}{l}{0.688/0.047} &  & \multicolumn{1}{l}{\textbf{0.694/0.020}} & \multicolumn{1}{l}{\textbf{0.725/0.014}} &  & \multicolumn{1}{l}{\textbf{0.683/0.021}} & \multicolumn{1}{l}{\textbf{0.725/0.094}} &  & \multicolumn{1}{l}{\textbf{0.617/0.013}} & \multicolumn{1}{l}{\textbf{0.752/0.005}} &  & \multicolumn{1}{l}{\textbf{0.792/0.038}} & \multicolumn{1}{l}{\textbf{0.808/0.035}} \\

WISER(Z) w/o $\vl_{cns}$&{0.593/0.041} & \multicolumn{1}{l} {\textbf{0.745/0.047}} &  & \multicolumn{1}{l}{0.664/0.013} & \multicolumn{1}{l}{0.675/0.024} &  & \multicolumn{1}{l}{0.568/0.027} & \multicolumn{1}{l}{0.603/0.082} &  & \multicolumn{1}{l}{0.510/0.079} & \multicolumn{1}{l}{0.551/0.114} &  & \multicolumn{1}{l}{0.736/0.017} & \multicolumn{1}{l}{0.754/0.021} \\

WISER(Z) w/o \{$\vl_{cns}, \vl_{embed}$\}& 0.545/0.032 & \multicolumn{1}{l}{0.483/0.023} &  & \multicolumn{1}{l}{0.617/0.018} & \multicolumn{1}{l}{0.601/0.027} &  & \multicolumn{1}{l}{0.487/0.024} & \multicolumn{1}{l}{0.540/0.045} &  & \multicolumn{1}{l}{0.450/0.050} & \multicolumn{1}{l}{0.422/0.026} &  & \multicolumn{1}{l}{0.417/ 0.058} & \multicolumn{1}{l}{0.520/0.013} \\
\hline
\end{tabular}

}
\label{tab:ablation_auroc_auprc}
\end{table*}
\section{Comparison with self supervised learning methods}
We have further compared our method with other self supervised learning approaches. Based on prior literature~\cite{alsaggaf2024improving}, we use Gaussian noise based perturbation of the genomic samples for the domain-invariant representations. On the augmented data, we apply various SSL methods like SimCLR~\cite{simclr} and Barlow Twins~\cite{zbontar2021barlow} on CODE-AE. The results (mean/std. over 5-fold cross validation) is provided in Table \ref{tab:comarision_ssl}.

\begin{table*}[h!]
\centering
\caption{Performance comparison of predicted patient response using AUROC and AUPRC metrics of our proposed method \textbf(WISER) with other self supervised learning approaches. Data related to clinical relapse is used for all the evaluations. The result is noted in the form of (mean / std) where the score has been obtained over five fold cross validation. On an average, our method outperforms others baselines on all the drugs for at least one metric. The best performer is highlighted in \textbf{bold}.}
\resizebox{\linewidth}{!}{
\begin{tabular}{l|lrlrrlrrlrrlrr}
\hline
\textbf{Methods} & \multicolumn{2}{c}{5-Fluorouracil} &  & \multicolumn{2}{c}{Temozolomide} &  & \multicolumn{2}{c}{Sorafenib} &  & \multicolumn{2}{c}{Gemcitabine} &  & \multicolumn{2}{c}{Cisplatin} \\
\hline
 & AUROC & \multicolumn{1}{l}{AUPRC} &  & \multicolumn{1}{l}{AUROC} & \multicolumn{1}{l}{AUPRC} &  & \multicolumn{1}{l}{AUROC} & \multicolumn{1}{l}{AUPRC} &  & \multicolumn{1}{l}{AUROC} & \multicolumn{1}{l}{AUPRC} &  & \multicolumn{1}{l}{AUROC} & \multicolumn{1}{l}{AUPRC} \\
 \hline
\textbf{WISER} &{0.715/0.036} & \multicolumn{1}{l}{	0.741/0.023} &  & \multicolumn{1}{l}{\textbf{0.760/0.006}} & \multicolumn{1}{l}{\textbf{0.786/0.019}} &  & \multicolumn{1}{l}{\textbf{0.727/0.007}} & \multicolumn{1}{l}{\textbf{0.728/0.024}} &  & \multicolumn{1}{l}{\textbf{0.649/0.037}} & \multicolumn{1}{l}{\textbf{0.752/0.002}} &  & \multicolumn{1}{l}{\textbf{0.851/0.007}} & \multicolumn{1}{l}{\textbf{0.861/0.020}} \\

CODE-AE & \textbf{0.868/0.030} & \multicolumn{1}{l}{0.740/0.006} &  & \multicolumn{1}{l}{0.751/0.017} & \multicolumn{1}{l}{0.762/0.001} &  & \multicolumn{1}{l}{0.631/0.020} & \multicolumn{1}{l}{0.705/0.062} &  & \multicolumn{1}{l}{0.594/0.016} & \multicolumn{1}{l}{0.751/0.006} &  & \multicolumn{1}{l}{0.652/0.071} & \multicolumn{1}{l}{0.743/0.011} \\

CODE-AE + SIMCLR & 0.663/0.051 & \multicolumn{1}{l}{0.699/0.129} &  & \multicolumn{1}{l}{0.707/0.007} & \multicolumn{1}{l}{0.733/0.024} &  & \multicolumn{1}{l}{0.479/0.024} & \multicolumn{1}{l}{0.610/0.027} &  & \multicolumn{1}{l}{0.490/0.029} & \multicolumn{1}{l}{0.609/0.027} &  & \multicolumn{1}{l}{0.469/0.05} & \multicolumn{1}{l}{0.518/0.070} \\

CODE-AE + Barlow Twins & 0.747/0.029 & \multicolumn{1}{l}{\textbf{0.767/0.048}} &  & \multicolumn{1}{l}{0.680/0.015} & \multicolumn{1}{l}{0.681/0.036} &  & \multicolumn{1}{l}{0.569/0.029} & \multicolumn{1}{l}{0.576/0.033} &  & \multicolumn{1}{l}{0.621/0.052} & \multicolumn{1}{l}{0.621/0.036} &  & \multicolumn{1}{l}{0.670/0.116} & \multicolumn{1}{l}{0.702/0.028} \\
\hline
\end{tabular}

}
\label{tab:comarision_ssl}
\end{table*}

\section {Generalization on other drugs and datasets}
 To establish the generalizability of proposed method in two aspects - (1) on unknown drugs (unseen during representation learning) and (2) on a different dataset, we conducted similar experiments on the PDTC breast cancer dataset~\cite{bruna2016biobank} (32 samples per drug), on drugs unused in TCGA. The results(mean/std. over 5-fold cross validation) are shown in Table \ref{tab:pdtc_results}.

\begin{table*}[h!]
\centering
\caption{Performance comparison of predicted response using AUROC and AUPRC metrics of our proposed method \textbf(WISER) on PDTC dataset. The result is noted in the form of (mean / std) where the score has been obtained over five fold cross validation. On an average, our method outperforms others baselines on two out of three drugs. The best performer is highlighted in \textbf{bold}.}
\resizebox{0.8\linewidth}{!}{
\begin{tabular}{l|lrlrrlrr}
\hline
\textbf{Methods} & \multicolumn{2}{c}{Az628} &  & \multicolumn{2}{c}{Gefitinib} &  & \multicolumn{2}{c}{Axitinib}  \\
\hline
 & AUROC & \multicolumn{1}{l}{AUPRC} &  & \multicolumn{1}{l}{AUROC} & \multicolumn{1}{l}{AUPRC} &  & \multicolumn{1}{l}{AUROC} & \multicolumn{1}{l}{AUPRC} \\
 \hline
\textbf{WISER} &{\textbf{0.792/0.069}} & \multicolumn{1}{l}{\textbf{0.789/0.029}} &  & \multicolumn{1}{l}{\textbf{0.700/0.025}} & \multicolumn{1}{l}{\textbf{0.793/0.031}} &  & \multicolumn{1}{l}{\textbf{0.864/0.011}} & \multicolumn{1}{l}{0.836/0.037} \\

CODE-AE &{0.754/0.097} & \multicolumn{1}{l}{0.679/0.148} &  & \multicolumn{1}{l}{0.613/0.037} & \multicolumn{1}{l}{0.778/0.009} &  & \multicolumn{1}{l}{0.840/0.049} & \multicolumn{1}{l}{0.762/0.033} \\
Velodrome &{0.513/0.015} & \multicolumn{1}{l}{0.625/0.064} &  & \multicolumn{1}{l}{0.446/0.091} & \multicolumn{1}{l}{0.495/0.052} &  & \multicolumn{1}{l}{0.786/0.041} & \multicolumn{1}{l}{\textbf{0.841/0.014}} \\

\hline
\end{tabular}

}
\label{tab:pdtc_results}
\end{table*}

\end{document}